\documentclass[11pt]{article}

\usepackage[final]{acl}

\usepackage{times}
\usepackage{latexsym}

\usepackage[T1]{fontenc}

\usepackage[utf8]{inputenc}

\usepackage{microtype}

\usepackage{inconsolata}

\usepackage{graphicx}

\title{BOSCH: Black-Box Binary Optimization for Short-Context Attention-Head Selection in LLMs}

\author{Abbas Ghaddar\thanks{Equal contribution}\hspace{3mm}Ivan Kobyzev\footnotemark[1]\hspace{3mm}Boxing Chen\hspace{3mm}Yufei Cui \\
Huawei Noah’s Ark Lab, Montreal Research Center, Canada\\
{\footnotesize\texttt{\{abbas.ghaddar,ivan.kobyzev,boxing.chen,yufei.cui\}@huawei.com}}}

\usepackage{amssymb}
\usepackage{pifont}

\newcommand{\bert}{BERT}
\newcommand{\bosch}{\textsc{BOSCH}}
\newcommand{\dcam}{\textsc{DCAM}}
\newcommand{\apl}{\textsc{APL}}
\newcommand{\qada}{\textsc{QAdA}}
\newcommand{\razor}{\textsc{Razor}}
\newcommand{\fisher}{\textsc{Fisher}}
\newcommand{\proxyattn}{\textsc{Proxy}}

\newcommand{\mightmention}[1]{}
\newcommand{\problem}[1]{\textcolor{red}{$\star$}}
\newcommand{\answer}[1]{\textcolor{blue}{$\#$}}
\newcommand{\todoreview}[1]{\textcolor{green}{$@$}}

\usepackage{subcaption}
\usepackage{rotating,csquotes}
\usepackage{xcolor}
\usepackage{multirow}
\usepackage{framed}
\usepackage{booktabs}
\usepackage{amsmath}
\usepackage{upgreek}
\usepackage{amsfonts}

\usepackage[most]{tcolorbox}

\newtcbox{\mybox}[1][]{enhanced, colframe=blue, colback=blue!15, 
	frame style={opacity=0.25}, interior style={opacity=0.25}, 
	nobeforeafter, tcbox raise base, shrink tight, extrude by=1mm, #1}

\usepackage{pgfplots}
\usepackage{array}
\usepackage{xcolor}
\usetikzlibrary{shapes.multipart}

\usepackage{subcaption}
\usepackage{rotating,csquotes}
\usepackage{xcolor}
\usepackage{multirow}
\usepackage{framed}
\usepackage{booktabs}
\usepackage{amsmath}
\usepackage{upgreek}
\usepackage{amsfonts}
\usepackage{arydshln}

\usepackage{amsmath,amssymb,mathtools}

\usepackage{algorithm}
\usepackage[noend]{algpseudocode} 

\usepackage{enumitem}

\usepackage{float}

\usepackage{algorithm}
\usepackage{algpseudocode} 

\usepackage{rotating}     
\usepackage{subcaption}   
\usepackage{booktabs}
\usepackage{caption}

\makeatletter
\AtBeginDocument{%
}
\makeatother

\usepackage{placeins} 
\begin{document}
\maketitle
\begin{abstract}

Post-training hybridization of large language models (LLMs) often replaces quadratic self-attention with sliding-window attention (SWA) to reduce KV cache usage and improve latency. Existing hybridization schemes are typically defined either at the layer level (e.g., interleaving) or at the head level via static rankings from local to global. Layer-level schemes ignore that local and global dependencies are routed through heads within the same layer, while static head-level rankings suffer from \emph{entanglement}: a head’s local/global behavior can change after hybridization. We propose \bosch{}, \emph{Black-box Binary Optimization for Short-context Head Selection}, a training-free method that formulates the problem as a Large Neighborhood Search and decomposes it into three subproblems: (i) \emph{layer-importance detection} via small-budget black-box probes, (ii) \emph{adaptive per-layer SWA-ratio assignment} based on these sensitivities, and (iii) \emph{grouped head-level optimization} within ratio buckets. Extensive experiments on 4 LLMs ranging from 1.7B to 30B parameters, across 4 SWA ratios, show that \bosch{} consistently outperforms layer-level heuristics and 6 strong static head-level methods, with larger gains at higher SWA ratios. Under continual pretraining, \bosch{} recover original long-context performance faster and to a higher level. Analysis of the selected heads reveals substantial turnover for \bosch{} across different SWA ratios, underscoring the importance of performing head-level selection for each target ratio rather than relying on fixed locality rankings. 

\end{abstract}

\section{Introduction}

Transformer self-attention~\cite{vaswani2017attention} is the core of the token-mixing mechanism in state-of-the-art Large Language Models (LLMs)~\cite{dubey2024llama,team2025kimi,yang2025qwen3}. Due to its quadratic complexity, there has been a line of research focused on building hybrid LLMs that combine self-attention with more efficient alternatives, such as State Space Models (SSMs)~\cite{mamba2,yang2025gated} or Sliding Window Attention (SWA)~\cite{Beltagy2020LongformerTL}. 

Hybridization is either performed by pretraining from scratch~\cite{openai2025gptoss120bgptoss20bmodel,li2025minimax,dong2025hymba} or via post-training hybridization~\cite{mambainllama,yang2025zebra,Gu2025JetNemotronEL}. In this paper, we focus on SWA post-training hybridization of a pretrained Transformer self-attention LLM to enable efficient long-context handling. We choose SWA not only for its constant time and memory footprint, but also for its zero-shot compatibility with self-attention, which enables zero-shot hybridization and requires only minimal training for near-complete performance recovery. Given a target ratio, the main challenge lies in defining the hybridization scheme, namely which components to replace and at what granularity: layer or head level.

Previous works on LLM hybridization~\cite{arxiv23_mistrial,gptoss,mambainllama,yang2025zebra} has primarily relied on rule-based heuristics at the layer level, such as layer interleaving or begin–middle–end (BME). More recent works~\cite{zhang-2024-draft,Gu2025JetNemotronEL} have reported improved results by deploying search-based algorithms, largely enabled by the limited search space induced by the relatively small number of layers in LLMs. However, layer-level granularity is misaligned with how transformers actually route information~\cite{clark2019bert}. Local and global dependencies are handled by different attention heads within the same layer~\cite{Olsson2022IncontextLA, Wu2024RetrievalHM, Tang2024RazorAttentionEK}. Consequently, flipping an entire layer from global to local attention can remove critical global information, leading to performance degradation.

Identifying long-context attention heads in Transformers has been an active research direction for many years, with applications in model interpretability~\cite{clark-2019-bert,pascual2021telling}, weight pruning~\cite{kwon2022fast}, KV-cache compression~\cite{Tang2024RazorAttentionEK}, attention sparsification~\cite{wang2025proxyattn}, and online SWA hybridization~\cite{Donhauser2025UnveilingSO}. In the context of SWA hybridization, these static methods rank heads from local to global and then convert the most local heads according to a given ratio. However, this approach suffers from an entanglement problem: the local–global behavior of heads estimated before hybridization may change after hybridization, leading to suboptimal performance.
 
A straightforward solution would be to apply a search-based algorithm directly at the head level. However, this is computationally prohibitive: modern LLMs expose hundreds to low thousands of heads, making brute-force search infeasible, and black-box optimization algorithms quickly stall because each evaluation is expensive and the probability that a single random flip improves the objective drops rapidly as dimensionality grows~\cite{ShanWang2010, Frazier2018Tutorial}. Even robust black-box methods such as mesh-adaptive direct search (MADS)~\cite{AudetDennisMADS2006} work best when each subproblem involves only tens of variables; beyond that, the number of evaluations required to find consistent improvements grows too quickly for practical budgets.

To address these issues, we formulate SWA head selection as a Large Neighborhood Search (LNS)~\cite{ShawLNS1998} problem and decompose it into 3 subproblems, making it feasible to perform the search under a realistic evaluation budget. We propose \bosch{}, a black-box binary optimization search-based method that \textit{(i)} constructs promising neighborhoods by ranking layers according to their localization sensitivity, \textit{(ii)} assigns adaptive per-layer SWA ratios, and \textit{(iii)} jointly optimizes the heads within each neighborhood under the target SWA ratio to produce the final SWA head-level hybridization scheme. 

We conduct extensive SWA hybridization experiments on 4 models ranging from 1.7B to 30B parameters, across 4 SWA ratios, comparing against 3 layer-level and 6 head-level baseline methods. Results on 6 \emph{needle-in-a-haystack} (NIAH) and 30 long-context QA tasks from the LongBench~\cite{bai2024longbench} benchmark show that our \bosch{} systematically outperforms both layer-level and static head-level prior methods, with the performance gap becoming more significant at higher SWA ratios. In addition, we show that \bosch{} recovers performance after continual training both faster and to a higher level than best existing methods. Our analysis reveals a correlation between method performance and the pairwise similarity of the heads selected by these methods. Moreover, we observe significant sets of heads that appear in the \bosch{} SWA head set at smaller ratios (e.g., 0.5) but not at higher ratios, indicating that mitigating entanglement is crucial for performance, thereby justifying \bosch{} superior performance.

\begin{figure*}[!ht]
    \centering
    \includegraphics[width=1.0\textwidth]{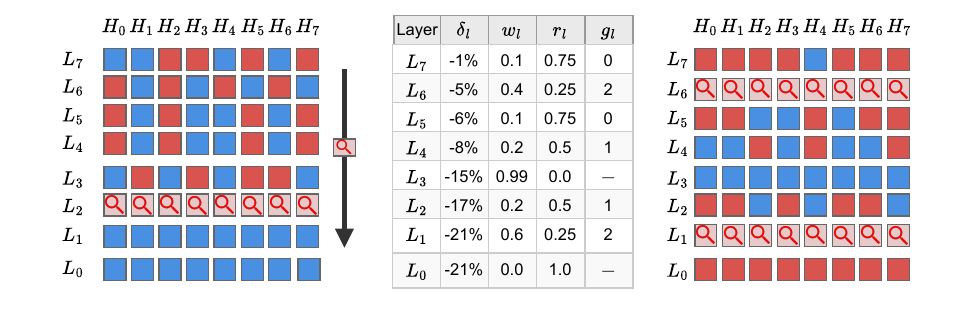}
    \caption{Illustrative application of \textbf{BOSCH} to a Transformer with $L{=}8$ layers and $H{=}8$ heads per layer, targeting $\rho\!=\!0.5\%$ SWA heads. \textbf{Left (Stage 1: layer-importance detection):} each row is a layer; blue squares are self-attention heads ($z{=}1$), red squares are SWA heads ($z{=}0$); light-red squares with a loupe indicate the layer(s) currently scored via black-box optimization, from top to bottom. \textbf{Middle (Stage 2: adaptive ratio assignment):} for each layer $\ell$ we compute the relative drop $\delta_\ell$ (from the original model), normalize it to an importance weight $w_\ell\!\in[0,1]$, and map it to a per-layer SWA ratio $r_\ell$ while respecting the global budget. \textbf{Right (Stage 3: multi-layer head selection under adaptive ratios):} layers are bucketed by ratio into groups $g_\ell$ and optimized jointly from more SWA-tolerant to less tolerant; the example highlights searching layers with $g_\ell{=}2$ using $r_\ell{=}0.25$, yielding the final head mask $z$. "$-$" indicates that either entire layer is full-attention or SWA ( no optimization is applied).}
    \label{fig:main_integral}

\end{figure*}

\section{Methodology}
\label{sec:Methodology}

\subsection{Problem Formalization}
\label{sec:formalization}

Let $\mathcal{M}$ be a pretrained LLM with $L$ layers and $H$ heads per layer. Denote the total number of heads as $N=LH$ and flatten heads to a single index $i\in\{1,\dots, N\}$. 
Let $z\in\{0,1\}^N$ be a binary mask, where $z_i=1$ corresponds to the global causal self-attention and $z_i=0$ to the sliding-window attention (SWA). Finally, denote $\mathcal{L}:(\mathcal{M},z,\mathcal{D})\to\mathbb{R}$ to be a loss evaluated on a calibration set $\mathcal{D}$. 
We formulate the SWA head selection as the following constrained binary black-box minimization problem: 
\begin{equation}
\label{eq:problem}
   \begin{aligned}
   \min_{z\in\{0,1\}^N}\  \mathcal{L}(\mathcal{M},z,\mathcal{D}), \quad \\
\quad\text{s.t.}\quad
\frac{1}{N}\sum_{i=1}^{N}\big(1-z_i\big)=  \rho ,
\end{aligned} 
\end{equation}
where $\rho \in[0,1]$ is the target ratio of SWA heads.

Additionally, if the model uses Grouped-Query Attention (GQA)~\cite{ainslie2023gqa}, we enforce that all heads sharing a KV group take the same decision, i.e., we impose $z_i=z_j$ for any pair $(i,j)$ of heads belonging to the same group. 
 This restriction is meaningful because in GQA the keys and values are shared per group. Switching a single head to SWA while other heads in its group remain full-attention does not reduce the KV cache size; memory savings occur only when the entire KV group switches.

\subsection{\bosch{}}
\label{sec:BOSCH}

In this section, we design a loss function for the SWA hybridization problem and introduce a tailored hybridization pipeline, \bosch{}, short for \underline{\textbf{B}}lack-Box Binary \underline{\textbf{O}}ptimization for \underline{\textbf{S}}hort-\underline{\textbf{C}}ontext Attention-Head \underline{\textbf{S}}election.

Let \(\mathcal{S}(\mathcal{M},z,\mathcal{D})\) denote the performance score (higher the better) of the model $\mathcal{M}$ on the calibration set \(\mathcal{D}\) where attention heads are localized according to the decision mask $z$.
To normalize the score to be from 0 to 1 in a meaningful way, we define two anchors: $a = \mathcal{S}(\mathcal{M},\{0\}^N,\mathcal{D})$, the performance of the total SWA model, and  $b = (1+\gamma)\mathcal{S}(\mathcal{M},\{1\}^N,\mathcal{D})$, the performance of the original full-attention model. We introduce a small factor \(\gamma>0\) to ensure a positive normalization span and reserve headroom for performance improvements beyond the original model.
Then we calculate the normalized performance score as:
\begin{equation}
\label{eq:score}
 \widehat{\mathcal{S}}(\mathcal{M},z,\mathcal{D}) \;=\; \frac{\mathcal{S}(\mathcal{M},z,\mathcal{D}) - a}{\,b-a\,} 
\end{equation}
leading to the design of our loss function:
\begin{equation}
\label{eq:loss}
\scalebox{0.85}{$
 \mathcal{L}(\mathcal{M},z,\mathcal{D},\rho)
= - \widehat{\mathcal{S}}(\mathcal{M},z,\mathcal{D}) \;+\; \alpha\,\big(\rho(z)- \rho\big)^2,
$}
\end{equation}
where $\rho(z)=\tfrac{1}{N}\sum_{i=1}^N (1-z_i)$ and $\rho$ is the target budget for the ratio of SWA heads. $\alpha>0$ trades off validation performance against adherence to the target ratio.

Directly applying off-the-shelf black-box binary optimizers to solve problem in Equation~\ref{eq:problem} with the loss from Equation~\ref{eq:loss} is impractical for two main reasons. First, evaluating the loss is expensive: each call to \(\mathcal{S}(\cdot)\) requires a LLM forward pass and typically takes several seconds, so only a small number of iterations is feasible, which severely limits search and leads to a poor solution.
Second, the search space is large. The number of binary variables is \(N{=}LH\), which is in the range from a few hundred to a few thousand for billion-scale LLM. 
State-of-the-art black-box neighborhood methods like MADS~\cite{AudetDennisMADS2006} explore $O(N)$ one-bit neighbors per poll, while the probability that any single bit flip yields improvement scales like $\sim\!1/N$. As a result, the number of evaluations required to find improvements grows rapidly, and evaluation budgets become infeasible already beyond roughly 50 variables. In practice, toolkits like NOMAD~\cite{nomad4paper} restrict themselves to subproblems with less than 50 variables, which severely constrains search efficiency and prevents effective global exploration.

Following a Large Neighborhood Search~\cite{ShawLNS1998} approach over the binary decision vector \(z\), we split the optimization into two complementary subproblems: (i) neighborhood construction, 
 where we identify promising subsets and per-layer budgets, and (ii) neighborhood optimization, where we jointly refine assignments within those neighborhoods. The full procedure comprises three stages, described below.

\subsubsection{Layer importance detection}
\label{sec:stage1}We first assess each layer’s sensitivity to attention-head localization by iterating from the topmost layer to the bottommost layer. For each layer $\ell$, we keep the current binary mask $z$ fixed for all layers except for the current layer and run a small-budget black-box search to convert exactly $\lceil\rho H \rceil$ heads in layer $\ell$ to SWA, maximizing the performance score $\mathcal{S}$. Then we include the found head configuration to update the binary mask $z$ and proceed to the layer $\ell-1$, so each decision is made on the updated model with previously localized upper layers.
During this iteration, for every step we record the resulting best score and collect these values into
$s_\text{best} \in \mathbb{R}^L$, which guides subsequent stages.  Figure~\ref{fig:main_integral} (left) illustrates the search for $l\!=\!2$ with upper layers already localized. Algorithm~\ref{alg:stage1_flat} in Appendix~\ref{app:algorithms} gives the exact routine.

\subsubsection{Adaptive ratio assignment}
\label{sec:stage2}
Given the per-layer scores \(s_{\text{best}}\) computed in Stage~1, we compute the performance drop per layer $\delta$ from the original full-attention model. Using these values we recompute the relative drop from layer to layer and rescale them to weights \(w_\ell\in [0,1]\), where lower \(w_\ell\) means easier to localize. 
Layers are then sorted and bucketed into a small number of groups that map to coarse localization ratios (the target fraction of heads to convert to SWA in that layer). We reconcile these initial assignments with the global budget by shifting layers between adjacent buckets: we raise “easier” layers until the total number of localized heads matches the target, and, if needed, lower the “harder” ones. The outcome is a vector of per-layer adaptive ratios $r_\ell$ that allocate where localization is safest under the budget; these ratios serve as quotas for the within-layer head selection in Stage 3.
See Figure~\ref{fig:main_integral} (middle) for an illustrative example and Algorithm~\ref{alg:stage2-fixed-buckets-aligned-diff-simplified} for detailed procedure.

\subsubsection{Multi-layer head selection under adaptive ratios}
\label{sec:stage3}
Given the per-layer adaptive ratios \(\{r_\ell\}\), we group layers that share the same ratio and process groups from more localizable to less (largest to smallest $r$). For each group, we jointly optimize the binary head decisions in its layers while keeping previously processed groups fixed. 
We run a small-budget black-box optimization over the concatenated heads' indices within the group to convert exactly $\lceil r_\ell H \rceil$ heads to SWA, where $r_\ell$ is the ratio for the current group. 
The resulting assignment is committed to the global mask \(z\) before proceeding to the next group. See Figure~\ref{fig:main_integral} (right) for an example, where we first localized layers in groups~0 and 1 with $r \ge 0.5$ and doing the search for heads in group~2 with $r=0.25$. The exact procedure is detailed in  Algorithm~\ref{alg:stage3_flat} in \autoref{app:algorithms}.

\begin{table*}[!th]
\centering
\resizebox{\textwidth}{!}{
\begin{tabular}{lcccccccccccccccc}
\toprule
\multirow{2}{*}{\textbf{Method}} &
\multicolumn{4}{c}{\bf $\rho\!=\!0.25$} &
\multicolumn{4}{c}{\bf $\rho\!=\!0.5$} &
\multicolumn{4}{c}{\bf $\rho\!=\!0.75$} &
\multicolumn{4}{c}{\bf $\rho\!=\!0.875$} \\
\cmidrule(lr){2-5}\cmidrule(lr){6-9}\cmidrule(lr){10-13}\cmidrule(lr){14-17}
 & \bf 1.7B & \bf 8B & \bf 14B & \bf 30B
 & \bf 1.7B & \bf 8B & \bf 14B & \bf 30B
 & \bf 1.7B & \bf 8B & \bf 14B & \bf 30B
 & \bf 1.7B & \bf 8B & \bf 14B & \bf 30B \\
\midrule
Original*
 & 92.7 & 99.1 & 99.6 & 99.4
 & 92.7 & 99.1 & 99.6 & 99.4
 & 92.7 & 99.1 & 99.6 & 99.4
 & 92.7 & 99.1 & 99.6 & 99.4 \\
\midrule
\multicolumn{17}{c}{\textit{\bf Layer-level Heuristics}} \\
\midrule
RAND
 & 66.9 & 45.9 & 79.0 & 79.6
 & 38.0 & 15.4 & 39.2 & 16.0
 & 12.7 & 12.8 & 17.1 & 15.0
 & 12.6 & 13.2 & 13.0 & 10.6 \\
BME
 & 40.0 & 30.8 & 41.3 & 47.5
 & 11.5 & 12.4 & 11.8 & 18.9
 & 11.1 & 12.2 & 11.9 & 12.0
 & 11.2 & 12.7 & 12.2 & 10.5 \\
INTR
 & 74.9 & 72.9 & 41.1 & 69.1
 & 19.1 & 19.0 & 14.1 & 12.8
 & 12.8 & 12.7 & 11.3 & 22.8
 & 12.5 & 12.6 & 11.4 & 11.3 \\
\midrule
\multicolumn{17}{c}{\textit{\bf Head-level Static Methods}} \\
\midrule
\dcam{}
 & 80.7 & 95.8 & 96.9 & 92.7
 & 12.3 & 84.4 & 87.0 & 81.2
 & 10.6 & 22.6 & 30.2 & 41.2
 & 10.6 & 13.2 & 13.2 & 9.5 \\
\apl{}
 & 22.9 & \underline{98.0} & \underline{99.0} & 89.1
 & 13.5 & 67.6 & 91.3 & 50.1
 & 10.6 & 13.0 & 14.1 & 10.4
 & 12.2 & 12.1 & 12.9 & 10.0 \\
\proxyattn{}
 & 50.9 & 97.8 & 97.3 & 88.3
 & 23.8 & 71.6 & 68.6 & 76.9
 & 11.3 & 31.6 & 42.7 & 29.2
 & 11.7 & 13.9 & 13.8 & 10.2 \\
\qada{}
 & 83.5 & 95.2 & 83.9 & 80.8
 & 60.2 & 75.9 & 70.8 & 61.8
 & 38.9 & 46.1 & 19.1 & 34.9
 & 16.5 & 13.0 & 17.9 & 10.8 \\
\razor{}
 & 85.5 & 94.1 & 98.9 & 85.0
 & 64.8 & 82.4 & 88.0 & 78.4
 & 47.8 & \underline{64.9} & 71.3 & 37.2
 & \underline{21.9} & \underline{33.9} & \underline{46.9} & 11.6 \\
\fisher{}
 & 87.2 & 94.2 & 98.8 & 92.8
 & \underline{76.4} & 89.3 & \underline{93.7} & 81.5
 & 49.4 & 63.4 & \underline{71.6} & \underline{47.1}
 & 10.8 & 29.0 & 39.9 & 11.1 \\
\midrule
\multicolumn{17}{c}{\textit{\bf Ours}} \\
\midrule
\bosch{}
 & \textbf{91.8} & \textbf{98.9} & \textbf{99.2} & \textbf{97.5}
 & \textbf{78.3} & \textbf{90.3} & \textbf{94.0} & \textbf{86.3}
 & \textbf{58.0} & \textbf{72.7} & \textbf{83.6} & \textbf{50.2}
 & \textbf{30.3} & \textbf{42.5} & \textbf{47.2} & \textbf{26.9} \\

\hspace{3mm}\textit{-single}
 & 86.9 & 91.0 & 96.3 & 76.3
 & 62.8 & 88.9 & 87.9 & 58.3
 & 27.9 & 41.9 & 31.2 & 26.9
 & 19.9 & 12.3 & 19.7 & 21.0 \\

\hspace{3mm}\textit{-multi}
 & 89.5 & 97.6 & 97.2 & 85.8
 & 71.2 & 60.0 & 88.8 & 72.4
 & \underline{57.3} & 48.5 & 60.1 & 31.5
 & 19.6 & 18.4 & 36.4 & \underline{23.7} \\

\hspace{3mm}\textit{-layer}
 & \underline{90.0} & 93.2 & 97.8 & \underline{96.1}
 & 69.4 & \underline{89.7} & 92.2 & \underline{85.0}
 & 49.4 & 40.6 & 56.4 & 40.3
 & 12.3 & 13.0 & 33.5 & 23.1 \\
\bottomrule
\end{tabular} %
}
\caption{Zero-shot average scores on the NIAH benchmark for SWA hybridization methods across 4 Qwen3 models (1.7B-30B) under 4 SWA $\rho$ ratios. The highest and second-highest scores for each model under each ratio are highlighted in bold and underline, respectively. *For readability purposes, we repeat the original model’s performance in each column (SWA hybridization is not applied to the original model).}
\label{tab:zs_main_niah}
\end{table*}

\begin{table*}[!th]
\centering
\resizebox{\textwidth}{!}{
\begin{tabular}{lcccccccccccccccc}
\toprule
\multirow{2}{*}{\textbf{Method}} &
\multicolumn{4}{c}{\bf $\rho\!=\!0.25$} &
\multicolumn{4}{c}{\bf $\rho\!=\!0.5$} &
\multicolumn{4}{c}{\bf $\rho\!=\!0.75$} &
\multicolumn{4}{c}{\bf $\rho\!=\!0.875$} \\
\cmidrule(lr){2-5}\cmidrule(lr){6-9}\cmidrule(lr){10-13}\cmidrule(lr){14-17}
 & \bf 1.7B & \bf 8B & \bf 14B & \bf 30B
 & \bf 1.7B & \bf 8B & \bf 14B & \bf 30B
 & \bf 1.7B & \bf 8B & \bf 14B & \bf 30B
 & \bf 1.7B & \bf 8B & \bf 14B & \bf 30B \\
\midrule
Original*
 & 45.4 & 57.1 & 58.4 & 54.9
 & 45.4 & 57.1 & 58.4 & 54.9
 & 45.4 & 57.1 & 58.4 & 54.9
 & 45.4 & 57.1 & 58.4 & 54.9 \\
\midrule
\multicolumn{17}{c}{\textit{\bf Layer-level Heuristics}} \\
\midrule
RAND
 & 33.7 & 40.2 & 48.2 & 40.2
 & 28.5 & 31.5 & 37.6 & 18.3
 & 23.3 & 24.1 & 32.8 & 15.6
 & \underline{23.1} & 24.3 & 31.4 & 13.8 \\
BME
 & 31.5 & 31.0 & 32.6 & 28.2
 & 21.1 & 24.4 & 15.9 & 16.7
 & 20.7 & 28.9 & 29.1 & 12.3
 & 20.9 & 26.5 & 30.2 & 14.0 \\
INTR
 & 36.0 & 43.3 & 43.4 & 42.3
 & 28.2 & 34.2 & 25.0 & 25.2
 & 19.8 & 29.1 & 30.9 & 14.9
 & 19.7 & 23.5 & 30.4 & 13.1 \\
\midrule
\multicolumn{17}{c}{\textit{\bf Head-level Static Methods}} \\
\midrule
\dcam{}
 & 32.5 & 43.0 & 52.5 & 40.8
 & 22.5 & 33.0 & 42.3 & 31.1
 & 18.4 & 26.0 & 33.6 & \underline{21.1}
 & 17.8 & 24.0 & 31.1 & 10.6 \\
\apl{}
 & 36.8 & \underline{51.1} & 53.2 & 41.8
 & 25.9 & 34.7 & 43.6 & 28.3
 & 16.3 & 25.3 & 30.0 & 13.3
 & 16.6 & 23.7 & 29.4 & 11.6 \\
\proxyattn{}
& 34.4 & 50.5 & 49.8 & 38.4
& 24.3 & 33.7 & 39.7 & 27.0
& 19.7 & 29.3 & 32.0 & 17.2
& 18.3 & 27.3 & 29.2 & 12.0 \\
\qada{}
 & 36.4 & 40.1 & 43.9 & 35.1
 & 28.1 & 30.0 & 36.4 & 27.0
 & \underline{25.0} & 21.2 & 31.2 & 20.4
 & 21.7 & 23.3 & 28.1 & 12.9 \\
\razor{}
 & 35.1 & 41.2 & 52.2 & 34.5
 & 27.9 & 29.2 & 45.6 & 30.5
 & 19.8 & 25.7 & \underline{35.9} & 18.8
 & 19.8 & 26.7 & \underline{33.0} & 14.3 \\
\fisher{}
 & 36.1 & 40.9 & \underline{55.3} & \underline{44.2}
 & \underline{29.7} & 34.7 & 39.6 & \underline{31.8}
 & 21.5 & 25.8 & 32.6 & 20.9
 & 19.5 & 23.6 & 30.6 & 13.8 \\
\midrule
\multicolumn{17}{c}{\textit{\bf Ours}} \\
\midrule

\bosch{}
 & \textbf{38.0} & \textbf{52.2} & \textbf{56.2} & \textbf{46.2}
 & \textbf{32.1} & \textbf{41.8} & \textbf{47.0} & \textbf{36.0}
 & \textbf{26.0} & \textbf{31.6} & \textbf{38.0} & \textbf{24.8}
 & \textbf{23.6} & \textbf{28.6} & \textbf{36.1} & \textbf{19.3} \\

\hspace{3mm}\textit{-single}
 & 36.3 & 45.5 & 50.3 & 34.0
 & 26.3 & \underline{36.4} & 36.7 & 27.4
 & 19.4 & 26.2 & 32.5 & 17.9
 & 18.5 & \underline{27.4} & 25.4 & 14.1 \\

\hspace{3mm}\textit{-multi}
 & 36.0 & 46.7 & 51.1 & 40.8
 & 29.5 & 34.1 & 38.0 & 28.2
 & 21.8 & 26.6 & 35.6 & 17.4
 & 20.3 & 26.9 & 26.2 & 14.7 \\

\hspace{3mm}\textit{-layer}
 & \underline{37.1} & 47.7 & 52.3 & 39.9
 & 27.4 & 36.3 & \underline{46.8} & 30.8
 & 24.7 & \underline{30.4} & 35.4 & 20.1
 & 19.1 & 22.0 & 32.2 & \underline{15.1} \\
\bottomrule
\end{tabular}
}
\caption{Zero-shot average scores on the LongBench benchmark for SWA hybridization methods across 4 Qwen3 models under 4 SWA $\rho$ ratios. Notations follow those used in \autoref{tab:zs_main_niah}.}
\label{tab:zs_main_longbench}
\end{table*}

\section{Experimental Setting}
\label{sec:Experiments}

In this section, we summarize the backbone models, implementation details, baseline methods from prior work, and evaluation protocols. A more detailed description of our experimental setting is provided in Appendix~\ref{app:Experimental Setting}.

\subsection{Models}
We conducted our main experiments on the Qwen3~\cite{yang2025qwen3} family of models, ranging from 1.7B to 30B parameters. We selected Qwen3 because its base variants achieve state-of-the-art performance among open-weight models of same category, and have a native support for 32k sequence length.

\subsection{SWA Configuration}
We conduct experiments in which $\rho\!\in\!\{0.25, 0.5, 0.75, 0.875\}$ of the attention heads, groups, or layers is converted to SWA. All reported results using SWA are with a window size of 1024 (32 times smaller than the model's maximum sequence length) unless otherwise specified.

\subsection{Calibration Data}
Our calibration set $\mathcal{D}$ comprises synthetic \emph{needle-in-a-haystack} (NIAH) examples~\cite{needle}. We compute $\mathcal{S}(\mathcal{M}, z, \mathcal{D})$ by running a prefill-only forward pass (no decoding) and, for each example, checking whether all answer tokens are predicted correctly. $\mathcal{S}(\mathcal{M}, z, \mathcal{D})$ is then the mean accuracy across examples:

\begin{equation}
\mathcal{S}(\mathcal{M}, z, \mathcal{D})
= \frac{1}{|\mathcal{D}|}\sum_{(x,y)\in\mathcal{D}}
\mathbf{1}\!\left[\hat{y}=\!y\right],
\end{equation}

where $\hat{y}$ are the model’s prefill predictions for the answer tokens. We use NIAH-style data because it offers a length-controllable probe of long-range associative \emph{recall} while minimizing train–test memorization, due to the random nature of the NIAH data-generation process. 

\subsection{Baselines}

We compare \bosch{} with baselines we constructed from prior works on long-context attention-head identification: \dcam~\cite{clark2019bert}, \apl~\cite{pascual2021telling}, 
\fisher~\cite{kwon2022fast},
\qada~\cite{Donhauser2025UnveilingSO}, 
\razor~\cite{Tang2024RazorAttentionEK},  and \proxyattn~\cite{wang2025proxyattn}. Also, we compare against the widely used \textit{interleave} (INTR) and begin-middle-end (BME) layer-level selection heuristics~\cite{mambainllama,Yang2025ZebraLlamaTE}, as well as a \textit{random} (RAND) layer selection baseline.\footnote{Results are averaged over three runs.}  Finally, we report the results of 3 \bosch{} ablations: directly using the output of stage 1 (\textit{B-single}); using the output of stage 1 when grouping multiple layers rather than single layers (\textit{B-multi}); and running stage 1 at layer-level granularity instead of head-level (\textit{B-layer}), such that all layers fit in a single run.  

\subsection{Evaluation Benchmarks}
We evaluate methods on two benchmark suites for long-context evaluation and report the average scores on each benchmark. We define the \textbf{NIAH} average as the mean performance on 6 Needle-in-a-Haystack tasks with context lengths ranging from 4k to 32k from \textsc{RULER}~\cite{hsieh2024ruler}. The \textbf{LongBench} average is the unweighted average score across 6 long-context QA task categories (29 tasks) from LongBench~\cite{bai2024longbench}, and one math reasoning task represented by GSM8K~\cite{cobbe2021training}. Tables in \autoref{app:Results} contain detailed per-task results for the experiments conducted in the next section.

\section{Results and Analysis}
\label{sec:Results and Analysis}

\subsection{Zero-Shot Results}
\label{sec:Zero-Shot Results}

\autoref{tab:zs_main_niah} and \autoref{tab:zs_main_longbench} report the zero-shot performance after applying SWA hybridization with 3 layer-level heuristics, 6 head-level static methods, as well as our \bosch{} and its 3 variants on the NIAH and LongBench benchmarks, respectively.

First, we notice that on both benchmarks, \bosch{} consistently outperforms all baselines under every SWA ratio and model size. Among head-level static methods, \fisher{} and \razor{} are the strongest competitors, but they remain clearly behind \bosch{}. Earlier BERT-oriented head-locality detection methods such as \dcam{} and \apl{} perform substantially worse in this LLM setting, suggesting that these analysis methods do not transfer well to modern decoder-only architectures. 

Second, we observe that the gap between \bosch{} and the competitor methods tends to widen as the SWA ratio $\rho$ increases, indicating that our method is particularly robust when a larger fraction of attention heads is replaced. In contrast, there is no clear and consistent similar trend with respect to model size from 1.7B to 30B parameters, pointing to model-agnostic behavior for our method. Also, we notice that commonly used layer-level heuristics for post-training hybridization, such as BME and INTR, perform poorly overall. They often fall in the same range as the RAND baseline, highlighting the limitations of simple structural heuristics. 

Third, the ablations that only use the stage-one scoring (\textit{B-single}) underperform the three-stage \bosch{} pipeline by a large margin. However, expanding stage-one selection to more layers (\textit{B-multi}) narrows the gap and approaches the performance of layer-level black-box search (\textit{B-layer}). While these two variants still systematically fall behind \bosch{}, they achieve competitive results with the best head-level static methods such as \fisher{} and \razor{}. This indicates the importance of mitigating entanglement effects either at the head level (\textit{B-multi}, \bosch{}) or the layer level (\textit{B-layer}), which our black-box search framework is able to do compared to static methods.

\subsection{Cross-method Heads Analysis}
\label{sec:Cross-Method Head Analysis}

\begin{figure}[!h]
    \centering
    \includegraphics[scale=0.8]{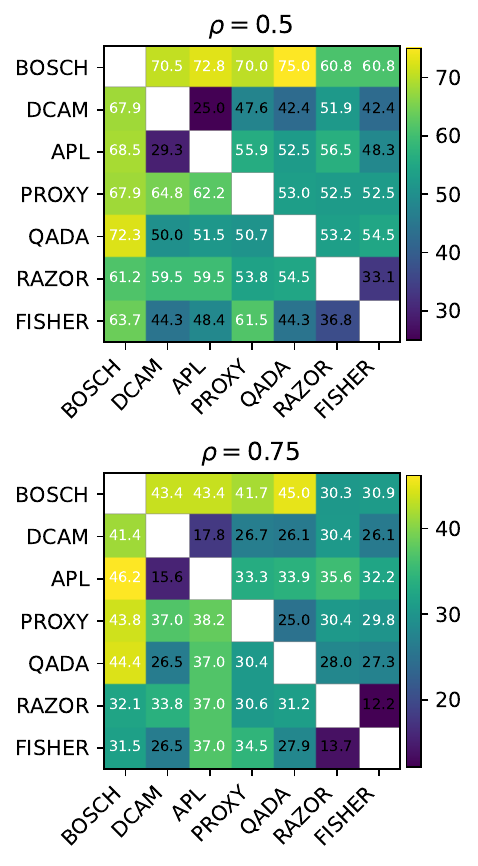}
    \caption{Jaccard distance between SWA heads selected by seven methods for $\rho = 0.25$ (upper plot) and $\rho = 0.5$ (lower plot). The distances for 8B and 14B models are shown in the lower and upper triangles, respectively.}
    \label{fig:heatmap_main}
\end{figure}

\autoref{fig:heatmap_main} shows the Jaccard distance heatmap between SWA-selected heads for 6 head-level static methods and our \bosch{} under $\rho \in \{0.5, 0.75\}$. 

Due to space constraints, we use the lower and upper triangles of each matrix to show results for the 8B and 14B models, respectively. Statistics for the remaining $\rho$ values and models are illustrated in \autoref{fig:heatmap_app_1.7b_30b} in \autoref{app:Results}. \bosch{} exhibits relatively large Jaccard distances to all static methods, indicating that it selects a distinctly different set of SWA heads compared to these methods. Interestingly, we observe a correlation between pairwise distances and method performance: methods that achieve similar NIAH/LongBench scores tend to share more heads (i.e., have lower distances). For example, \fisher{} and \razor{} form a tighter pair than the other static methods (and are also the closest to \bosch{} among them). In contrast, \dcam{} and \apl{} are close to each other yet remain far from the high-performing group, while \proxyattn{} and \qada{} each select their own relatively distinct sets of heads. This may suggest that black-box, search-based methods have the flexibility to discover interactions between heads, allowing them to identify a distinct and robust set of local attention heads.

\begin{figure*}[!th]
    \centering
    \includegraphics[width=\linewidth]{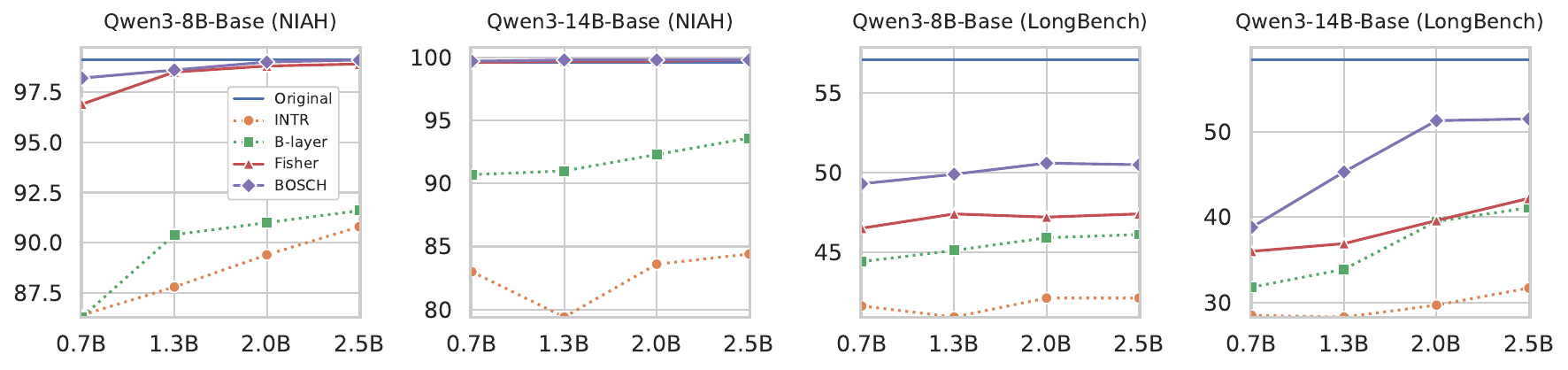}

    \caption{NIAH (first 2 plots) and LongBench (last 2 plots) performances (y-axis) for Qwen3-8B-Base and Qwen3-14B-Base models as a function of intermediate checkpoints during continual pretraining on 2.5B tokens (x-axis). We report scores for two layer-level and two head-level SWA hybridization methods using $\rho\!=\!0.75$. The performance of the full-attention original models (without additional training) is shown as a straight line with no markers.}
    \label{fig:cont_train}
\end{figure*}

\subsection{\bosch{} Heads Analysis}
\label{sec:Bosch Head Analysis}

To further validate this, we compute the fraction of heads $T$ (the turnover ratio) that are selected by \bosch{} with a small $\rho_s$ (e.g., $0.5$) but not selected by \bosch{} with a larger $\rho_l$ (e.g., $0.75$). We also define the geometric turnover rate $\tilde{T}$, which normalizes $T$ by its maximum possible value $T_{\max}(\rho_s, \rho_l)$ given only the two $\rho$ (e.g., $T_{\max}(0.5, 0.75)\!=\! 0.5$).

\begin{table}[!th]
\centering
\resizebox{\columnwidth}{!}{
\begin{tabular}{l|cc|cc|cc}
\toprule
\multirow{2}{*}{\textbf{Model}} &
\multicolumn{2}{c|}{$0.25 \rightarrow 0.5$} &
\multicolumn{2}{c|}{$0.5 \rightarrow 0.75$} &
\multicolumn{2}{c}{$0.75 \rightarrow 0.875$} \\
& $T$ & $\tilde{T}$ & $T$ & $\tilde{T}$ & $T$ & $\tilde{T}$ \\
\midrule
1.7B & 30\% & 30\% & 15\% & 30\% & 6\% & 36\% \\
8B   & 26\% & 26\% & 15\% & 29\% & 7\% & 44\% \\
14B  & 28\% & 28\% & 16\% & 31\% & 5\% & 31\% \\
30B  & 29\% & 29\% & 16\% & 31\% & 6\% & 34\% \\
\bottomrule
\end{tabular}
}
\caption{ Raw turnover $T$ and geometrically normalized turnover $\tilde{T}$
between \bosch{} local-head sets at 2 adjacent ratios $\rho$. All values are reported as percentages.}
\label{tab:turnover_bosch}
\end{table}

\autoref{tab:turnover_bosch} shows both statistics on 4 Qwen3 models, with 3 adjacent $\rho$ pairs. We observe that, across all models and $\rho$ pairs, there is a non-negligible number of heads present at $\rho_s$ but not at $\rho_l$. Additionally, for the same $\rho$ pairs, all models exhibit roughly the same range of $T$ rates. Although $T$ tends to decrease as $\rho$ increases (which is expected, since $T_{\max}$ decreases), it remains consistently within the defined range of 26\% (lowest) to 44\% (highest), with most $\tilde{T}$ values around 30\%. These trends suggest that \textit{starting over} at each target $\rho$ to anneal the impact of local head entanglement is crucial for performance.\footnote{We further confirm this finding with additional experiments in \autoref{app:BOSCH Head Analysis}.} While \bosch{} must restart at each target $\rho$, making it more expensive than static methods that run local-head detection and ranking only once, this added cost is justified by its consistently higher performance.

\subsection{Continual Training Results}
\label{sec:Continual Training Results}

Although SWA hybridization yields substantial latency reductions,\footnote{See \autoref{app:Latency and Memory Analysis} for a latency and memory analysis.} it also leads to siginficant  performance degradation, particularly at higher $\rho$. We therefore conduct continual pretraining experiments with $\rho\!=\!0.75$, which provides a favorable efficiency–quality trade-off, to evaluate how much of the original performance can be recovered after hybridization.

\autoref{fig:cont_train} presents intermediate NIAH and LongBench scores for Qwen3-8B-Base and Qwen3-14B-Base during continual pretraining on 2.5B tokens for 4 methods: INTR and \textit{B-layer} (layer-level), \bosch{} and the strongest head-level static baseline \fisher{}. The training setup is detailed in \autoref{app:Continual Pretraining}, and full results, including those for 1.7 and 30B models, are reported in \autoref{app:Results}.

On the one hand, we observe that head-level methods can substantially recover the original model performance better than layer-level methods. Both \bosch{} and \fisher{} nearly close the NIAH gap and significantly reduce the LongBench gap after 2.5B additional tokens, whereas layer-level methods remain clearly underperforming. However, our black-box layer-level variant \textit{B-layer} consistently outperforms the INTR heuristic, suggesting that black-box search is beneficial even when restricted to layer granularity. 

On the other hand, \bosch{} retains a clear advantage over \fisher{} after continual pretraining, especially on LongBench, indicating that better head masks at initialization translate into better downstream generalization even when extra training is allowed. Interstingly, even with with a high $\rho\!=\!0.75$, 2.5B tokens are sufficient to almost fully recover long-context recall on NIAH and to substantially narrow the gap on LongBench. This is encouraging when compared to architecture-level hybridization approaches such as MambaInLlama or Jet-Nemotron, which require orders of magnitude more training tokens and pipeline training. Our results suggest that post-hoc SWA hybridization, combined with a modest amount of continual pretraining, can be an efficient and practical alternative for deploying long-context–efficient variants of existing LLMs.

\section{Related Work}
There has been a surge of hybrid models that combine Transformer layers with more efficient modules like linear SSM/RNN blocks \cite{mamba2, Lieber2024JambaAH, dong2024hymba, MiniMax2025MiniMax01SF, qwen3-next-blog} or sliding-window attention (SWA) \cite{gptoss, arxiv23_mistrial}. These models are trained from scratch and typically require substantial pretraining compute.  
To reduce compute, continual-training hybridization have been proposed \cite{mambainllama, yang2025zebra, lu2021rw,lu2025regla}: starting from a pre-trained transformer it replaces full-attention with more-efficient blocks post-hoc.  In these works, the placement decision is at the layer level via a fixed interleaving schedule. 

Recently, Jet-Nemotron \cite{Gu2025JetNemotronEL} approaches hybridization via neural architecture search: starting from a pre-trained full-attention Transformer, the authors freeze MLP weights, search over placements of residual full-attention layers and candidate linear-attention blocks, and use beam search on a calibration set to select configurations for a target hybridization ratio. Our method can replace beam search with an advanced black-box search algorithm and apply a head-level linear-attention hybridization warm-up instead of layer-level mixing. 

A complementary line of work targets training-free head-level KV-cache reduction. RazorAttention~\cite{Tang2024RazorAttentionEK} is a static method that identifies attention heads important for long-context processing (so called “retrieval heads") using a synthetic repeated-token probe and truncates distant tokens for the rest. \citet{Donhauser2025UnveilingSO} propose a dynamic per-token detector that classify long context-relevant heads based on the attention mass condensation in the window near the boundary. Unlike these detection-based techniques, \bosch\  directly optimizes a binary head mask under explicit budgets, taking into account the inter-head dependency and the entangled behavior. 

Finally, post-training pruning removes attention heads without retraining: \citet{kwon2022fast} propose Fisher-guided structured pruning with mask rearrangement and per-layer output reconstruction, and ProxyAttn \cite{wang2025proxyattn} estimates block importance training-free by pooling scores from representative heads and allocating a dynamic sparsity budget. In contrast to these per-component, statistics-driven methods, our approach formulates selection as black-box binary optimization enabling joint optimization across heads.

\section{Conclusion}

In this work, we propose \bosch{}, a training-free black-box binary optimization framework for short-context attention-head selection for LLMs SWA post-training hybridization. We decompose the problem into 3 stages: layer-importance detection, adaptive per-layer SWA-ratio assignment, and grouped head-level search. \bosch{} consistently outperform layer-level heuristics and strong static head-level baselines on long-context benchmarks across multiple models sizes and SWA ratios. As future work, we plan to extend \bosch{} to other hybrid primitives incorporating SSM layers, where a key challenge is the current inability to perform zero-shot search. We also aim to adapt our framework to support Multi-Latent Attention~\cite{liu2024deepseek}, in which all attention heads are compressed into a single continuous vector.

\section*{Limitations}

Our experiments are conducted exclusively on the Qwen3 family in order to prioritize the systematicity of our evaluation. For instance, other model families~\cite{team2024gemma,arxiv23_llama2,dubey2024llama,abdin2024phi} either natively support relatively shorter maximum sequence lengths (e.g., 4k or 8k tokens), do not offer multiple model sizes, or are only available as instruction-tuned variants. Due to computational constraints, we limit our exploration of continual pretraining to a single data mixture and at most 2.5B additional tokens, and we do not experiment with models larger than 30B parameters. Finally, we do not study how SWA hybridization interacts with other efficiency techniques such as quantization, KV-cache compression, or weight pruning, leaving their compatibility and potential compounding effects for future work.

\section*{Acknowledgments}
We thank the anonymous reviewers for their insightful comments.

\bibliography{custom}

\appendix
\clearpage
\onecolumn

\section{\bosch{} Algorithms}
\label{app:algorithms}

We present algorithms for Stages 1–3 introduced in \autoref{sec:Methodology}. To this end, we slightly generalize the loss in Formula~\eqref{eq:loss} to accommodate the smaller‑cardinality subproblems arising in the neighborhood optimization steps. The general form of the loss is unchanged but we modify the penalty term.   
Let $N$ be the total number of heads and $\mathfrak{I} \subset \{1, \dots, N \}$ is a subset of heads. Define the loss as: 
\begin{equation}
\label{eq:loss_ind}
\begin{split}
\mathcal{L}(\mathcal{M},z,\mathcal{D},\bar\rho,\mathfrak{I})
&= -\,\widehat{\mathcal{S}}(\mathcal{M},z,\mathcal{D})
   \;+\; \alpha\big(\bar\rho_{\mathfrak{I}}(z)-\bar\rho\big)^2, \\
\text{where}\quad
\bar\rho_{\mathfrak{I}}(z)
&= \frac{1}{|\mathfrak{I}|}\sum_{i\in\mathfrak{I}}(1-z_i).
\end{split}
\end{equation}
As for the original loss,  $\widehat{\mathcal{S}}$ is the normalized score function from Formula~\ref{eq:score}, $\bar\rho$ is the target ratio of SWA heads for the subproblem we are solving, and $\alpha>0$ trades the validation performance against the adherence to the target ratio.

For Stage~1 we optimize at the layer granularity. For each layer $\ell$ the index set of heads is: 
\begin{equation}
    \mathfrak{I}_\ell= \{(\ell-1)\cdot H+1, \dots, (\ell-1)\cdot H+H \}. 
\end{equation}

\begin{algorithm}[H]
\caption{Stage 1}
\label{alg:stage1_flat}
\begin{algorithmic}[1]
\Require Model $\mathcal{M}$, calibration set $\mathcal{D}$, target ratio of SWA heads $\rho$
\Require Layers $L$, heads $H$, Scoring Function $\mathcal{S(\cdot)}$, Black-Box Optimizer BBO

\State $N \gets L H$;\quad $z \gets \mathbf{1}\in\{0,1\}^{N}$;\quad $s_{\mathrm{best}}\gets \mathbf{0} \in \mathbb{R}^L$ 
\For{$\ell = L,\ L{-}1,\ \dots,\ 1$} 
  \State $i_0 \gets (\ell-1)\cdot H + 1$;\quad $i_1 \gets i_0 + H $;\quad $\mathfrak{I}_\ell = \{i_0, \dots, i_1\}$  \Comment{indices of heads in the layer $\ell$}
  \State $\mathcal{U} \gets \{\,u\in\{0,1\}^{N}:\ u[k]=z[k] \quad \forall k \notin \mathfrak{I}_\ell ,\}$   \Comment{fix the mask outside of the layer $\ell$}
   \State $z \gets \min_{u\in\mathcal{U}} \mathcal{L}(\mathcal{M},u,\mathcal{D},\rho, \mathfrak{I}_\ell) $  \Comment{optimize the loss from Eq.~\ref{eq:loss_ind} with the BBO}
  \State $s_{\mathrm{best}}[\ell] \gets \mathcal{S}(\mathcal{M},z,\mathcal{D})$ 
\EndFor
\State \Return $s_{\mathrm{best}}$ 
\end{algorithmic}
\end{algorithm}

For Stage~3 we optimize for the heads based on the adaptive ratios. We explicitly form the index set by unifying the groups with the same ratio and use this ratio as the target budget for SWA heads. 

\begin{algorithm}[H]
\setcounter{algorithm}{2} 
\caption{Stage 3}
\label{alg:stage3_flat}
\begin{algorithmic}[1]
\Require Model $\mathcal{M}$, calibration set $\mathcal{D}$
\Require Layers $L$, heads $H$, per-layer adaptive ratios $\{r_\ell\}_{\ell=1}^{L}$
\Require Scoring Function $\mathcal{S(\cdot)}$, Black-Box Optimizer BBO
\State $N \gets L H$;\quad $z \gets \mathbf{1}\in\{0,1\}^{N}$ 
\State $M_\ell \gets \lceil r_\ell H\rceil$ for $\ell=1,\dots,L$
\State Sort layers by $r_\ell$ in descending order and split into groups $G_1,\dots,G_K $ with identical $r_\ell$.
\For{$g=1$ to $K$}
  \State $\mathfrak{I}_g \gets \bigcup_{\ell\in G_g}\ \{\ell H +1 ,\ \ell H{+}2,\ \dots,\ \ell H{+}H\}$ \Comment{indices of all heads in the current group}
  \State $\mathcal{U} \gets \Big\{u\in\{0,1\}^{N}:\  u[k]=z[k] \quad \forall k \notin \mathfrak{I}_g\}$ \Comment{fix the mask outside of the current group}
  \State $z \gets \min_{u\in\mathcal{U}} \mathcal{L}(\mathcal{M},u,\mathcal{D}, r_l, \mathfrak{I}_g) $  \Comment{optimize the loss from Eq.~\ref{eq:loss_ind} with the BBO}
\EndFor
\State \Return $z$
\end{algorithmic}
\end{algorithm}

\begin{algorithm}[H]
\setcounter{algorithm}{1} 
\caption{Stage 2}
\label{alg:stage2-fixed-buckets-aligned-diff-simplified}
\begin{algorithmic}[1]
\Require Layer scores $s_{\mathrm{best}}\in \mathbb{R}^L$, score of the full-attention model $s_{\text{orig}}$
\Require Target ratio of SWA heads $\rho$, layers $L$, heads per layer $H$
\Require Buckets $B$, partition of the unit interval $\{b_0<\cdots<b_{B-1}\}\subset[0,1]$ 
\Require Percentiles $p_{\mathrm{low}}{=}2.5$, $p_{\mathrm{high}}{=}97.5$
\State $N \gets L H$;\quad $M_{\text{target}} \gets \rho N$
\State $\delta_\ell \gets \dfrac{s_{\mathrm{best}}[\ell]-s_{\text{orig}}}{s_{\text{orig}}}$ for $\ell{=}1{:}L$;\quad $\delta_{L+1} \gets 0$ \Comment{compute absolute performance drop}
\State $d_\ell \gets \delta_\ell - \delta_{\ell+1}$ for $\ell{=}1{:}L$ \Comment{compute per layer relative performance drop}
\State $q_{\mathrm{lo}},q_{\mathrm{hi}} \gets \text{$p_{\mathrm{low}}$, $p_{\mathrm{high}}$ percentiles of }\{d_\ell\}$
\For{$\ell=0$ to $L{-}1$}
  \State $\tilde d_\ell \gets \min\{\max(d_\ell,\,q_{\mathrm{lo}}),\,q_{\mathrm{hi}}\}$ \Comment{clip to central percentiles}
  \State $w_\ell \gets \dfrac{\tilde d_\ell - q_{\mathrm{lo}}}{q_{\mathrm{hi}}-q_{\mathrm{lo}}}$ \Comment{compute normalized weights per layer}
\EndFor
\State Sort layers by $w_\ell$ in descending order and split into $B$ equal groups $G_0,\ldots,G_{B-1}$
\State Set $\texttt{rank}[\ell]\gets j$ for $\ell\in G_j$
\State $M_{\text{now}}\gets \sum_{\ell} b_{\texttt{rank}[\ell]} H$ 
\State $\Delta\gets M_{\text{target}}-M_{\text{now}}$ \Comment{head count gap; positive means more head can be localized}
\State $\texttt{easy}\gets$ layers sorted by $w_\ell$ ascending;\quad $\texttt{hard}\gets$ layers sorted by $w_\ell$ descending
\While{$\Delta \neq 0$}
  \If{$\Delta > 0$}
    \For{$\ell$ in $\texttt{easy}$}
      \If{$\texttt{rank}[\ell] < B{-}1$}
        \State $r\gets \texttt{rank}[\ell]$;\quad $\delta \gets H\,(b_{r+1}-b_{r})$
        \If{$\delta \le \Delta$}
          \State $\texttt{rank}[\ell]\gets r+1$;\quad $\Delta\gets \Delta - \delta$
          \If{$\Delta = 0$} \State \textbf{break} \EndIf
        \EndIf
      \EndIf
    \EndFor
  \Else
    \For{$\ell$ in $\texttt{hard}$}
      \If{$\texttt{rank}[\ell] > 0$}
        \State $r\gets \texttt{rank}[\ell]$;\quad $\delta \gets H\,(b_{r}-b_{r-1})$
        \If{$\delta \le -\Delta$}
          \State $\texttt{rank}[\ell]\gets r-1$;\quad $\Delta\gets \Delta + \delta$
          \If{$\Delta = 0$} \State \textbf{break} \EndIf
        \EndIf
      \EndIf
    \EndFor
  \EndIf
\EndWhile
\State $r_\ell \gets b_{\texttt{rank}[\ell]}$ for all $\ell$
\State \Return $\{r_\ell\}_{\ell=1}^{L}$ \Comment{adaptive ratios}
\end{algorithmic}
\end{algorithm}

\clearpage
\twocolumn

\section{Experimental Setting}
\label{app:Experimental Setting}

\subsection{Models}

\begin{table}[!th]
\centering
\footnotesize
\setlength{\tabcolsep}{4pt}
\resizebox{\columnwidth}{!}{%
\begin{tabular}{lccccc}
\toprule
\textbf{Model} & \textbf{\#P} & \textbf{\#L} & \textbf{\#QH} & \textbf{\#KVH} & \textbf{ML} \\
\midrule
Q3-1.7B    & 1.7  & 28 & 16 & 8 & 32k \\
Q3-8B      & 8.2  & 36 & 32 & 8 & 32k \\
Q3-14B     & 14.8 & 40 & 40 & 8 & 32k \\
Q3-30B-A3B & 30.5 & 48 & 32 & 4 & 40k \\
\bottomrule
\end{tabular}%
}
\caption{Qwen3-Base backbones used in our experiments. We report parameter count (\#P, in billions), number of layers (\#L), number of query attention heads (\#QH), number of key/value heads (\#KVH), and maximum length (ML, tokens).}
\label{tab:app_qwen3_base_singlecol}
\end{table}

\autoref{tab:app_qwen3_base_singlecol} summarizes the main characteristics of the Qwen3-Base models used in this study. All models use grouped-query attention (GQA) for efficient self-attention computation and natively support sequence lengths of at least 32k tokens (the 30B models support up to 40k tokens). The largest 30B model is a Mixture-of-Experts (MoE) model~\cite{shazeer2017outrageously} with only 3B parameters activated during inference, while all other models are dense. Therefore, it sometimes underperforms fully dense models (e.g., the 14B) in some settings.

\subsection{\bosch{} Implementation Details}
We use the MADS algorithm~\cite{AudetDennisMADS2006}, as implemented in the NOMAD toolkit~\cite{ledigabel2011nomad, nomad4paper}, as our binary black-box optimizer. Its surrogate quadratic-model search is only enabled for problems with at most 50 variables, which otherwise backoff to OnePlusOne~\cite{droste2002analysis} random variable selection. For example, with 50 variables we can group at most four layers (i.e., enforce $\max |G_g|\le 4$ in Algorithm~\ref{alg:stage3_flat}) for the 1.7B, 8B, and 14B Qwen3 models, each with 8 KV heads. If the number of layers that number, $G_g$ of Algorithm~\ref{alg:stage3_flat} is partitioned into smaller equally size subsets.

Let $\kappa$ denote the per-layer iteration budget, we  we use $\kappa\!=\!100$ in all experiments unless noted. In all experiments with \bosch{}, we allocate $\kappa$ iterations per layer across both \bosch{} stages and in the layer-level ablation (\textit{B-layer}). For instance, line~5 in Algorithm \ref{alg:stage1_flat} runs for $\kappa$ iterations each time it is called; line~7 in Algorithm \ref{alg:stage3_flat} runs for $\kappa\,|G_g|$ iterations for a group $G_g$; and the layer-level ablation runs for $\kappa\,L$ iterations. If, in any case, the number of possible candidates is less than or equal to the total iteration budget, we fall back to a brute-force algorithm that evaluates all candidates. Finally, we set $\alpha$ and $\gamma$ in \autoref{sec:BOSCH} to 100 and 0.2, respectively.

\bosch{} has two implementation modes: search and deploy. The search mode is used during optimization on the calibration set $\mathcal{D}$; for each forward pass and candidate mask $z$, hybridization is applied only at the locations indicated by $z$. The deploy mode is used after the final mask $z$ has been selected, for both inference and training. Both modes are implemented in PyTorch~\cite{paszke2019pytorch} on top of the Transformers library~\cite{wolf2020transformers} and are fully compatible with FlashAttention-2~\cite{daoflashattention} for efficient computation of both self-attention and SWA.

\noindent\textbf{Search mode}
We load the original self-attention model and, for each pass over $\mathcal{D}$ with a given mask $z$, handle any layer $\ell$ that requires hybrid SA/SWA heads as follows. After computing the $QKV$ projections, we partition the heads into SA and SWA subsets according to $z$. We then run the SA and SWA kernels in parallel (on two streams) and concatenate their outputs along the head dimension to form the attention output tensor. Before applying the output projection, we permute the rows of $W_o$ to match the 
\([\,\text{SA},\ \text{SWA}\,]\) head ordering of the concatenated tensor, and then apply $W_o$. This procedure requires loading the model once and allows $z$ to be updated efficiently after each full pass over $\mathcal{D}$. When grouped-query attention (GQA) is used, slicing respects KV-group boundaries: all heads in a KV group share the same decision.

\noindent\textbf{Deploy mode}
Given the final mask $z$, we materialize the per-layer hybrid modules once at model initialization. For each hybrid layer $\ell$, we slice the $QKV$ projection weight matrices along the head dimension into self-attention (SA) and sliding-window attention (SWA) subsets according to $z$, and permute the rows of $W_o$ to match the \([\,\text{SA},\ \text{SWA}\,]\) head ordering. At inference or training time, we compute the SA and SWA attention outputs in parallel (e.g., via separate kernel launches), concatenate them along the head dimension, and apply the output projection with $W_o$. When using GQA, slicing respects KV-group boundaries so that all heads within a KV group share the same decision.

\clearpage
\onecolumn   

\subsection{Baselines}
\label{app:baselines}

We define 6 training-free long-context head-selection methods built on prior works. These baselines were not originally designed for selecting SWA heads for model hybridization, but rather for other related tasks such as model interpretability~\cite{clark-2019-bert,pascual2021telling}, weight pruning~\cite{kwon2022fast}, KV-cache compression~\cite{Tang2024RazorAttentionEK}, attention sparsification~\cite{wang2025proxyattn}, and online SWA hybridization~\cite{Donhauser2025UnveilingSO}. Let denote by \(\alpha_{\ell,h}(t,j)\in[0,1]\) the attention probability at layer \(\ell\in\{1,\dots,L\}\) and head \(h\in\{1,\dots,H\}\) from source position \(j\) to target position \(t\) (causal, \(j\le t\)). Lags are denoted by \(d=t-j\), where all lag are self-included (\(d{=}0\)). With GQA, we compute each KV-group’s score as the mean over the heads in that group, as it consistently outperformed max pooling across methods in preliminary experiments. Each baseline yields a per-head score \(s_{\ell,h}\) where larger values indicate more local unless noted.  Given a target SWA ratio $\rho$, we rank heads (or groups) by from local to global and apply SWA to the top $\lceil \rho N \rceil$ (setting $z_i=0$), leaving the rest as full self-attention ($z_i=1$). SWA window $W$ is set to the same value (e.g. 1024) used when experimenting with our \bosch{} method.

\subsubsection{Distance-Conditioned Attention Mass (\textsc{DCAM})}
\label{app:dcam}

\citet{clark-2019-bert} studied \bert{}~\cite{devlin2019bert} attention head specialization patterns by conditioning on relative token distance by quantifying how much mass fell on nearby versus distant tokens. We adapt their method to obtain a single locality score per head ($s_{\ell,h}$) by aggregating attention scores by lag and normalizing to a per-head histogram:

\begin{equation}
\label{eq:dcma_1}
    M_{\ell,h}(d)\;=\!\!\sum_{(x,t)\in\mathcal{D}}\alpha_{\ell,h}(t,\,t{-}d),\qquad
p_{\ell,h}(d)\;=\;\frac{M_{\ell,h}(d)}{\sum_{u\ge 0}M_{\ell,h}(u)},
\end{equation}

Given an SWA window $W$, we define the \dcam{} locality score as the cumulative mass inside the window:

\begin{equation}
\label{eq:dcma_2}
s_{\ell,h}\;=\;\sum_{d=0}^{W-1}p_{\ell,h}(d),
\end{equation}
where higher $s_{\ell,h}$ indicates a more local head. We compute $s_{\ell,h}$ on the calibration set $\mathcal{D}$ and rank heads in descending order.

\subsubsection{Answer-Centric Peak Lag (\apl)}
\label{app:apl}

\citet{pascual2021telling} analyze heads via distance-conditioned attention patterns to characterize local vs. global heads in \bert{}-like models. We adapt this idea to our NIAH calibration set $\mathcal{D}$ by anchoring on the last answer token and measuring how far each head’s peak attention lies from it.

For each example $(x,y)\!\in\!\mathcal{D}$, let $t^\star(x)$ be the last answer token and define the peak lag
\begin{equation}
j^\star_{\ell,h}(x)=\arg\max_{j\le t^\star(x)}\alpha_{\ell,h}\!\big(t^\star(x),j\big),\qquad
\lambda_{\ell,h}(x)=t^\star(x)-j^\star_{\ell,h}(x).
\end{equation}
We aggregate across examples and normalize by the SWA window $W$ to obtain a locality score (larger $s_{\ell,h}$ means more local head):

\begin{equation}
\bar\lambda_{\ell,h}=\operatorname{median}_{x\in\mathcal{D}}\lambda_{\ell,h}(x),\qquad
s_{\ell,h}=1-\min\!\left(1,\frac{\bar\lambda_{\ell,h}}{W}\right).
\end{equation}

\subsubsection{Proxy Attention (\proxyattn)}
\label{app:proxy}

\textsc{ProxyAttn}~\cite{wang2025proxyattn} is a training-free guided sparse-attention algorithm that pools a few representative \emph{proxy} heads to compute unified token/block importance scores and assigns head-specific dynamic sparsity budgets. 
We adapt the their budget estimator into a per-head locality score on $\mathcal{D}$. For each example $(x,y)\in\mathcal{D}$ and \emph{target position} $t$, partition the \emph{source positions} $j\!\in\!\{1,\dots,t\}$ into $n_B$ contiguous blocks of size $B$. Form a pooled, layer-wise proxy attention by taking the maximum over a small proxy set $\mathcal{P}_\ell\subseteq\{1,\dots,H\}$ (default $|\mathcal{P}_\ell|{=}1$):
\begin{equation}
\tilde{\alpha}_{\ell}(t,j)\;=\;\max_{h\in\mathcal{P}_\ell}\alpha_{\ell,h}(t,j),
\end{equation}

then we compute pooled block masses and a common ranking:

\begin{equation}
S_\ell(b,t)\;=\!\!\sum_{j\in \text{block }b}\tilde{\alpha}_\ell(t,j),\qquad \text{then sort } \{S_\ell(b,t)\}_{b=1}^{n_B}\ \text{in descending order.}
\end{equation}

For each head $h$, let $m_{\ell,h}(t)$ be the smallest number of top-ranked blocks whose own cumulative mass reaches a target fraction $\gamma\in(0,1]$:
\begin{equation}
m_{\ell,h}(t)\;=\;\min\Big\{m'\!:\ \sum_{b=1}^{m'}\ \sum_{j\in \text{block }b}\alpha_{\ell,h}(t,j)\ \ge\ \gamma \sum_{b=1}^{n_B}\ \sum_{j\in \text{block }b}\alpha_{\ell,h}(t,j)\Big\}.
\end{equation}
We define the per-row fractional budget $q_{\ell,h}(t)=\tfrac{m_{\ell,h}(t)}{n_B}$ and the head score:

\begin{equation}
s_{\ell,h}\;=\;\mathbb{E}_{(x,t)\in\mathcal{D}}\!\big[q_{\ell,h}(t)\big],
\end{equation}

where larger \(s_{\ell,h}\) means that the mass is concentrated in fewer nearby blocks and consequently is more local head. Therefore, we rank heads by ascending order.

\subsubsection{Query-ADaptive Attention Criterion (\qada)}
\label{app:qada}

\qada{}~\cite{Donhauser2025UnveilingSO} aims to estimate, for each token and head, how much attention mass lies inside a local window  without explicitly computing full-context scores. They do so by modeling the non-local (global) token scores with a second-moment Gaussian approximation. We adapt \qada{} to produce a single locality score per head only on our calibration set $\mathcal{D}$.
For a given example $(x,y)\!\in\!\mathcal{D}$, layer $\ell$, head $h$, and target position $t$, let $q_{\ell,h}(t)\in\mathbb{R}^{d_k}$ and $k_{\ell,h}(j)\in\mathbb{R}^{d_k}$ be the query and key vectors.
We define the \emph{local} window and its complement:
\begin{equation}
\mathcal{N}_W(t)=\{\,j:\ \max(1,t{-}W{+}1)\le j\le t\,\},\qquad
\mathcal{F}_W(t)=\{\,j:\ 1\le j\le t{-}W\,\}.
\end{equation}
then we compute the exact unnormalized local softmax mass:
\begin{equation}
A_{\text{local}}(t)\;=\;\sum_{j\in \mathcal{N}_W(t)} \exp\!\big(\tau\, q_{\ell,h}(t)^\top k_{\ell,h}(j)\big).
\end{equation}
To approximate the \emph{non-local} mass, we model keys in $\mathcal{F}_W(t)$ as Gaussian along the query direction using a small boundary buffer $\mathcal{B}_B(t)=\{t{-}W{-}B{+}1,\dots,t{-}W\}\subseteq \mathcal{F}_W(t)$ to estimate second moments:
\[
\hat\mu_{\ell,h}(t)=\frac{1}{|\mathcal{B}_B(t)|}\!\!\sum_{j\in\mathcal{B}_B(t)}\!k_{\ell,h}(j),\qquad
\hat\Sigma_{\ell,h}(t)=\frac{1}{|\mathcal{B}_B(t)|}\!\!\sum_{j\in\mathcal{B}_B(t)}\!\big(k_{\ell,h}(j)-\hat\mu_{\ell,h}(t)\big)\big(k_{\ell,h}(j)-\hat\mu_{\ell,h}(t)\big)^\top.
\]
Using the Gaussian MGF, the expected global contribution is:
\begin{equation}
\widehat{A}_{\text{global}}(t)\;\approx\;|\mathcal{F}_W(t)|\cdot
\exp\!\Big(\tau\,q_{\ell,h}(t)^\top \hat\mu_{\ell,h}(t)\;+\;\tfrac{1}{2}\tau^2\,q_{\ell,h}(t)^\top \hat\Sigma_{\ell,h}(t)\,q_{\ell,h}(t)\Big).
\end{equation}
We then form the query-adaptive \emph{local-mass fraction}:
\begin{equation}
\widehat{\pi}_{\ell,h}(t)\;=\;\frac{A_{\text{local}}(t)}{A_{\text{local}}(t)+\widehat{A}_{\text{global}}(t)}, 
\end{equation}
and aggregate over rows/examples on $\mathcal{D}$ to obtain a single head score (larger means more local):
\begin{equation}
s_{\ell,h}\;=\;\mathbb{E}_{(x,t)\in\mathcal{D}}\!\big[\widehat{\pi}_{\ell,h}(t)\big]\;\in[0,1].
\end{equation}

\subsubsection{Razor Attention (\razor)}
\label{app:razor}

Razor Attention~\cite{Tang2024RazorAttentionEK} is a training-free KV cache compression method that aims to detects retrieval (long-context) heads, thus only keeping them in KV cache during decoding phase.
Retrieval heads are detected via a synthetic repeated-token probe that isolates two patterns: \emph{echo} (attending to the previous identical token) and \emph{induction} (attending to the token that follows the current token in an earlier occurrence). The authors constructed an input of length $4K$ by repeating a random length-$K$ token block four times. For target positions $m\in\{K,\dots,4K{-}1\}$, they define the \emph{echo} and \emph{induction} source indices:

\begin{equation}
j_{\mathrm{echo}} = m-K, \qquad
j_{\mathrm{ind}}  = m-1-K \ \ (\text{valid for } m \ge K{+}1),
\end{equation}

and compute, for each layer $\ell$ and head $h$, the mean attention to these sources across the probe:

\begin{equation}
E_{\ell,h} \;=\; \mathbb{E}_m\!\big[\alpha_{\ell,h}(m, j_{\mathrm{echo}})\big], 
\qquad
I_{\ell,h} \;=\; \mathbb{E}_m\!\big[\alpha_{\ell,h}(m, j_{\mathrm{ind}})\big].
\end{equation}

We convert these to across-head $z$-scores and into locality scores:
\begin{equation}
s_{\ell,h} \;=\; \max\!\big(z(E_{\ell,h}),\, z(I_{\ell,h})\big),
\end{equation}
where smaller values indicate more local heads. We set $K$ to one quarter of the sequence length of examples in $D$, to align with other methods.

\subsubsection{Fisher-Weighted Locality (\fisher)}
\label{app:fisher}

\citet{kwon2022fast} propose a post-training structured pruning method that uses an empirical Fisher approximation of the loss to rank and select network components like attention heads and FFN blocks. We adpat this method by only considering attention heads and converting the Fisher-weighted attention mass into a per-head locality scores. Let $\alpha_{\ell,h}(t,j)$ be the attention probability from source $j$ to target $t$ (causal, $j\!\le\!t$). We compute token-level cross-entropy $\mathcal{L}$ and its gradient in Fisher-style saliency on attention probabilities:
\begin{equation}
\Phi_{\ell,h}(t,j)\;=\;\Big(\tfrac{\partial \mathcal{L}}{\partial \alpha_{\ell,h}(t,j)}\cdot \alpha_{\ell,h}(t,j)\Big)^{\!2}, 
\end{equation}
then we bin by lag $d{=}t{-}j$ and normalize to a histogram:
\begin{equation}
F_{\ell,h}(d)\;=\!\!\sum_{(x,t)\in\mathcal{D}}\Phi_{\ell,h}\!\big(t,\,t{-}d\big),\qquad
q_{\ell,h}(d)\;=\;\frac{F_{\ell,h}(d)}{\sum_{u\ge 0}F_{\ell,h}(u)}.
\end{equation}
Given an SWA window $W$, the \fisher{} \emph{locality} score is the Fisher-weighted mass inside the window:
\begin{equation}
s_{\ell,h}\;=\;\sum_{d=0}^{W-1} q_{\ell,h}(d)\,\in[0,1],
\end{equation}
where larger values indicate more \emph{local} heads.

\clearpage
\twocolumn

\subsection{Calibration dataset}

Our calibration set $\mathcal{D}$ contains 64 examples, each a 32k-token sequence, generated using in-house code. We verified that it has no content overlap with the NIAH test set used in the \textsc{RULER} benchmark~\cite{hsieh2024ruler} to ensure a fair evaluation. With our efficient \bosch{} search mode, a single $\mathcal{S}(\mathcal{M}, z, \mathcal{D})$ call takes 2.2 seconds on Qwen3-1.7B-Base and 4.8 seconds on Qwen3-30B-A3B-Base, on a single node equipped with 8 A800 GPUs. Consequently, running both stages of \bosch{} for a single configuration can take up to 4 hours on Qwen3-1.7B-Base and up to 14 hours on Qwen3-30B-A3B-Base, respectively. The same calibration dataset $\mathcal{D}$ is used to run baseline methods of \autoref{app:baselines}. In comparison, the static baseline methods require less then one hour regardless of the model size using the same compute resources as our \bosch{}. While \bosch{} introduces higher overhead than baseline methods, the search is performed offline once per (model, SWA ratio), so the cost is amortized across all subsequent deployments and inference runs.

\subsection{Continual Pretraining}
\label{app:Continual Pretraining}

To recover the full performance of the original self-attention model, we continually pre-train hybrid SWA models produced by our \bosch{} method or the baselines on a small set of subsampled 2.5B tokens of Prolong~\cite{gao2024train} data. We perform tokenization and pack examples into fixed-length 32k token sequences. Models were trained on 2 GPU nodes, each equipped with 8 NVIDIA A800 cards with 80GB of memory. To accelerate pretraining, we use Fully Sharded Data Parallel (FSDP)~\cite{zhao2023pytorch}, mixed-precision training~\citep{fp16}, FlashAttention-2~\cite{shahflashattention}, and Liger-Kernel~\cite{hsu2024liger}. We train all models on fully packed sequences of 32,768 tokens and set the maximum per-GPU batch size per model, ranging from 8 for the smallest Qwen3-1.7B-Base to 2 for the largest Qwen3-30B-A3B-Base. We further speed up training by adjusting the gradient-accumulation steps to achieve a total batch size of 2M tokens, as recommended by~\cite{tangrethinking,ghaddar2021jaber,ghaddar2022revisiting}. For all models, we use the AdamW optimizer~\cite{loshchilov2017decoupled} with a cosine learning-rate scheduler, an initial learning rate of 1e-5, and a 10\% warmup. Continual pretraining took approximately one day for Qwen3-1.7B-Base and 4 days for Qwen3-30B-A3B-Base. 

\subsection{Evaluation Benchmarks}

We evaluate models on 2 benchmarks focusing on long-context reasoning and associative recall. We did not consider language understanding and modeling tasks such as ARC~\cite{allenai:arc}, PIQA~\cite{bisk2020piqa}, and MMLU~\cite{hendrycksmeasuring}, as they are regarded as short-context tasks where the total sequence length is smaller than our SWA window size. 

For \textsc{RULER}~\cite{hsieh2024ruler}, we include three NIAH-single (\texttt{niah\_single\_\{1,2,3\}}) and three NIAH-multikey (\texttt{niah\_multikey\_\{1,2,3\}}) subtasks evaluated at 4k, 8k, 16k, and 32k context lengths. For each length, we report the unweighted average across the three NIAH-single subtasks and across the three NIAH-multikey subtasks. The final \textbf{NIAH} average score is the mean over the 6 subtasks scores.

For \textbf{LongBench}~\cite{bai2024longbench} (v1), we follow the authors’ categorizations: Single-document QA (MultiFieldQA-en, MultiFieldQA-zh, NarrativeQA, Qasper), Multi-document QA (2WikiMQA, HotpotQA, MuSiQue, DuReader), Summarization (GovReport, MultiNews, QMSum, VCSum, SAMSum), Few-shot learning (TREC, TriviaQA, LSHT), Synthetic Tasks (PassageRetrieval-en, PassageRetrieval-zh, PassageCount), and Code Completion (LCC, RepoBench-P). We add an extra category, which we call \textit{Math Reasoning}, consisting of a single task: GSM8K~\cite{cobbe2021training}. The \textbf{LongBench} score is the unweighted average over seven categories (the 6 LongBench categories plus Math Reasoning).

\section{Analysis}
\label{app:Analysis}

\subsection{\bosch{} Head Analysis}
\label{app:BOSCH  Head Analysis}

We conduct additional experiments to validate whether the findings of \autoref{sec:Bosch Head Analysis} on \bosch{} heads selection is correlated with model performance. Let $\mathcal{A}_{\rho}$ denote the set of selected heads at ratio $\rho$. For two small and larger ratios $\rho_s < \rho_l$, define $\mathcal{A}_{\rho_s}$ and $\mathcal{A}_{\rho_l}$ to be the heads selected by these two ratios, respectively. Let
\[
\Delta \mathcal{A}_{s,l} = \mathcal{A}_{\rho_s} \setminus \mathcal{A}_{\rho_l}
\]
be the set of heads that exist at ratio $\rho_s$ but not at $\rho_l$; using this set, we randomly select the same number of heads from $\mathcal{A}_{\rho_l}$ and replace them with the heads in $\Delta \mathcal{A}_{s,l}$, leading to $\mathcal{A}^\prime_{\rho_l}$.

\begin{table*}[!th]
\centering
\resizebox{\textwidth}{!}{
\begin{tabular}{l|cc|cc|cc||cc|cc|cc}
\toprule
\multirow{3}{*}{\textbf{Model}} &
\multicolumn{6}{c||}{\textbf{NIAH}} &
\multicolumn{6}{c}{\textbf{LongBench}} \\
\cmidrule(lr){2-7}\cmidrule(lr){8-13}
& \multicolumn{2}{c|}{$0.25 \rightarrow 0.5$} &
  \multicolumn{2}{c|}{$0.5 \rightarrow 0.75$} &
  \multicolumn{2}{c||}{$0.75 \rightarrow 0.875$} &
  \multicolumn{2}{c|}{$0.25 \rightarrow 0.5$} &
  \multicolumn{2}{c|}{$0.5 \rightarrow 0.75$} &
  \multicolumn{2}{c}{$0.75 \rightarrow 0.875$} \\
& $\mathcal{A}_{\rho_l}$ & $\mathcal{A}^\prime_{\rho_l}$ & $\mathcal{A}_{\rho_l}$ & $\mathcal{A}^\prime_{\rho_l}$ & $\mathcal{A}_{\rho_l}$ & $\mathcal{A}^\prime_{\rho_l}$ &
  $\mathcal{A}_{\rho_l}$ & $\mathcal{A}^\prime_{\rho_l}$ & $\mathcal{A}_{\rho_l}$ & $\mathcal{A}^\prime_{\rho_l}$ & $\mathcal{A}_{\rho_l}$ & $\mathcal{A}^\prime_{\rho_l}$ \\
\midrule
1.7B  & 78.3 & 57.1 & 58.0 & 47.1 & 30.3 & 19.7 & 32.1 & 27.4 & 26.0 & 23.0 & 23.6 & 17.0 \\
8B    & 90.3 & 79.5 & 72.7 & 68.7 & 42.5 & 20.7 & 41.8 & 34.3 & 31.6 & 29.2 & 28.6 & 24.3 \\
14B   & 94.0 & 87.9 & 83.6 & 73.9 & 47.2 & 41.1 & 47.0 & 40.4 & 38.0 & 34.0 & 36.1 & 29.9 \\
30B   & 86.3 & 67.2 & 50.2 & 36.8 & 26.9 & 15.7 & 36.0 & 28.4 & 24.8 & 22.3 & 19.3 & 14.7 \\
\bottomrule
\end{tabular}
}
\caption{Zero-shot average scores on the NIAH and LongBench benchmarks for \bosch{} across 4 Qwen3 models  under 3 SWA ratios $\rho_l\in\{0.5,0.75,0.875\}$. The left side of $\rightarrow$ corresponds to $\rho_s$, while the right side corresponds to $\rho_l$.
$\mathcal{A}^\prime_{\rho_l}$ indicates results obtained after replacing a randomly selected subset of heads at a given configuration (e.g., $\rho_l\!=\!0.75$) with heads that did not appear in a smaller configuration (e.g., $\rho_s\!=\!0.5$). Results in the $\mathcal{A}^\prime_{\rho_l}$ columns are averaged over 3 runs with different seeds. The $\mathcal{A}_{\rho_l}$ column shows the results for the originally selected SWA heads set.}
\label{tab:app_turnover_bosch}
\end{table*}

\begin{table*}[!th]
\centering
\begin{tabular}{lccccccccccccc}
\toprule
\multirow{2}{*}{\textbf{Method}} &
\multicolumn{4}{c}{\textbf{single}} &
\multicolumn{4}{c}{\textbf{multikey}} &
\multicolumn{4}{c}{\textbf{Avg.}} \\
\cmidrule(lr){2-5}\cmidrule(lr){6-9}\cmidrule(lr){10-13}
 & \bf 1.7B & \bf 8B & \bf 14B & \bf 30B
 & \bf 1.7B & \bf 8B & \bf 14B & \bf 30B
 & \bf 1.7B & \bf 8B & \bf 14B & \bf 30B \\
\midrule
\multicolumn{13}{c}{\textit{64k}} \\
\midrule
Original         & 47.9 & 49.3 & 49.1 & 49.3 & 23.5 & 44.7 & 45.9 & 45.7 & 35.7 & 47.0 & 47.5 & 47.5 \\
\hdashline
INTR             &  2.6 &  2.9 &  2.6 & 17.3 &  1.7 &  1.8 &  1.3 &  2.5 &  2.2 &  2.4 &  2.0 &  9.9 \\
\textit{B-layer} & 23.7 & 18.1 & 12.1 & 19.5 &  8.2 &  1.6 &  2.7 &  3.8 & 16.0 &  9.9 &  7.4 & 11.7 \\
\fisher{}        & 27.7 & 37.0 & 37.0 & 23.9 &  7.5 &  8.7 & 13.3 &  5.7 & 17.6 & 22.9 & 25.2 & 14.8 \\
\bosch{}         & \textbf{32.0} & \textbf{41.3} & \textbf{43.7} & \textbf{38.5} & \textbf{15.0} & \textbf{13.9} & \textbf{23.6} & \textbf{11.1} & \textbf{23.5} & \textbf{27.6} & \textbf{33.7} & \textbf{24.8} \\
\midrule
\multicolumn{13}{c}{\textit{128k}} \\
\midrule
Original         & 21.6 & 24.3 & 24.4 & 22.4 &  6.1 & 14.7 & 18.1 & 16.5 & 13.8 & 19.5 & 21.3 & 19.4 \\
\hdashline
INTR             &  1.5 &  1.7 &  1.5 &  6.1 &  1.0 &  1.1 &  0.9 &  1.5 &  1.3 &  1.4 &  1.2 &  3.8 \\
\textit{B-layer} &  4.5 &  9.9 &  2.2 &  8.7 &  2.3 &  1.2 &  1.1 &  1.5 &  3.4 &  5.6 &  1.7 &  5.1 \\
\fisher{}        &  6.9 &  9.1 & 15.3 &  8.8 &  1.8 &  1.1 &  6.4 &  3.1 &  4.4 & 5.1& 10.9 &  6.0 \\
\bosch{}         & \textbf{11.7} & \textbf{10.0} & \textbf{18.7} & \textbf{15.3} & \textbf{5.1} & \textbf{2.4} & \textbf{6.5} & \textbf{4.9} & \textbf{8.4} & \textbf{6.2} & \textbf{12.6} & \textbf{10.1} \\

\bottomrule
\end{tabular}
\caption{Detailed zero-shot performance on the NIAH benchmark at \textit{64k} and \textit{128k} for the original models and four SWA hybridization methods (with SWA ratio $\rho = 0.75$) applied to four Qwen3 models (1.7B, 8B, 14B, 30B). Zero-shot length extrapolation is performed using a YaRN factor of 2 and 4 for 64k and 128k, respectively. The \textbf{Avg.} block reports, for each model, the mean of its single and multikey scores. The highest score under each configuration is highlighted in bold.}
\label{tab:app_extrapol}
\end{table*}

\autoref{tab:app_turnover_bosch} shows the zero-shot average NIAH and LongBench performance for \bosch{} when using three randomly sampled $\mathcal{A}^\prime_{\rho_l}$ head sets based on $\mathcal{A}_{\rho_l}$ (mean of 3 runs with different $\mathcal{A}^\prime_{\rho_l}$). Results for $\mathcal{A}_{\rho_l}$ are reported for reference, while the detailed NIAH and LongBench scores are presented in \autoref{tab:app_turnover_niah} and \autoref{tab:app_turnover_longbench}, respectively. We notice a significant drop in performance across all models, benchmarks, and ratios when using $\mathcal{A}^\prime$ instead of $\mathcal{A}$. For instance, on NIAH, the gap varies between $6\%$ (14B, $\rho_s \in \{0.5, 0.875\}$) and $20\%$ (14B, $\rho_s = 0.5$; 8B, $\rho_s = 0.875$). These observations further indicate that an SWA hybridization method should not only rely on pre-hybridization head locality identification, but also explicitly account for the impact of head entanglement after hybridization.

\begin{figure*}[!th]
    \centering
    \includegraphics[width=\linewidth]{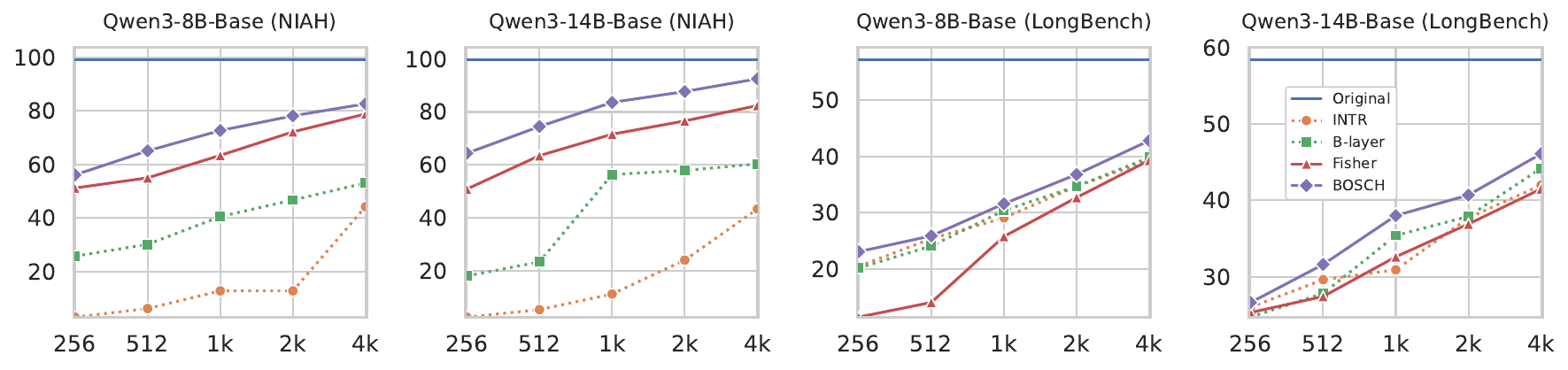}

    \caption{NIAH (first 2 plots) and LongBench (last 2 plots) zero-shot average performances (y-axis) for Qwen3-8B-Base and Qwen3-14B-Base models as a function of varying the SWA window size (x-axis). We report scores for two layer-level and two head-level SWA hybridization methods using $\rho\!=\!0.75$. The performance of the full-attention original models (without additional training) is shown as a straight line with no markers.}
    \label{fig:app_ws_8b_14b}
\end{figure*}

\subsection{Length Extrapolation Ablation}
\label{app:Length Extrapolation Ablation}
We evaluate on the NIAH benchmark using sequences that exceed each model’s native maximum context length to study how robust SWA hybridization methods are to longer sequence lengths. For all models, we perform zero-shot extrapolation of the context length by replacing the default RoPE positional encoding~\cite{su2024roformer} with the Yarn context window extension technique~\cite{pengyarn}. More precisely, we set the YaRN factor to 2.0 and 4.0 for inference sequence lengths of \textit{64k} and \textit{128k}, respectively, without any additional finetuning.

\autoref{tab:app_extrapol} reports NIAH zero-shot performance at \textit{64k} and \textit{128k} sequence lengths for the original model, as well as for the four SWA hybridization methods using with SWA ratio $\rho\!=\!0.75$. 
We find that \bosch{} remains the strongest SWA hybrid across all model sizes and both sequence lengths, consistently outperforming INTR, \textit{B-layer}, and \fisher{}. We also observe that the relative ranking among methods is largely stable compared to the 32k setting, while \bosch{}’s margin over layer-level heuristics becomes more significant as context length increases. This suggests that \bosch{} is not only effective within the native 32k window, but also transfers reasonably well under zero-shot length extrapolation to longer sequences such as 64k and 128k.

\subsection{SWA Windows Size Ablation}
\label{app:SWA Windows Size Ablation}

We conduct an ablation study by varying the SWA window size to better understand how the methods in our study generalize when evaluated with a window size that was not used during search or ranking. \autoref{fig:app_ws_8b_14b} shows the NIAH and LongBench zero-shot average scores of 4 methods for Qwen3-8B-Base and Qwen3-14B-Base when varying the SWA window size with values $\{256, 512, 1024, 2048, 4096\}$ (x-axis), while search and ranking are performed with a fixed window size of 1024 and SWA ratio $\rho\!=\!0.75$. Results for the Qwen3-1.7B-Base and Qwen3-30B-A3B-Base models, as well as detailed results for all models, are reported in \autoref{tab:app_ws_niah_075} and \autoref{tab:app_ws_longbench_075} for NIAH and LongBench, respectively.

Across all tested window sizes and models, \bosch{} systematically continues to outperform the remaining methods on both benchmarks. Moreover, the relative ranking of the baselines largely matches the one observed at the search window size ($W{=}1024$), with only few expections. In addition, we observe that increasing the window size generally leads to an  approximately monotonic performance improvements under most setting. This suggests that the quality of a head-selection scheme is fairly stable and predictable when the SWA windows size change at inference time evaluation, and that \bosch{}’s advantage does not limited to the window size used during search.

\subsection{Latency and Memory Analysis}
\label{app:Latency and Memory Analysis}

\begin{figure*}[!th]
    \centering
    \includegraphics[width=\linewidth]{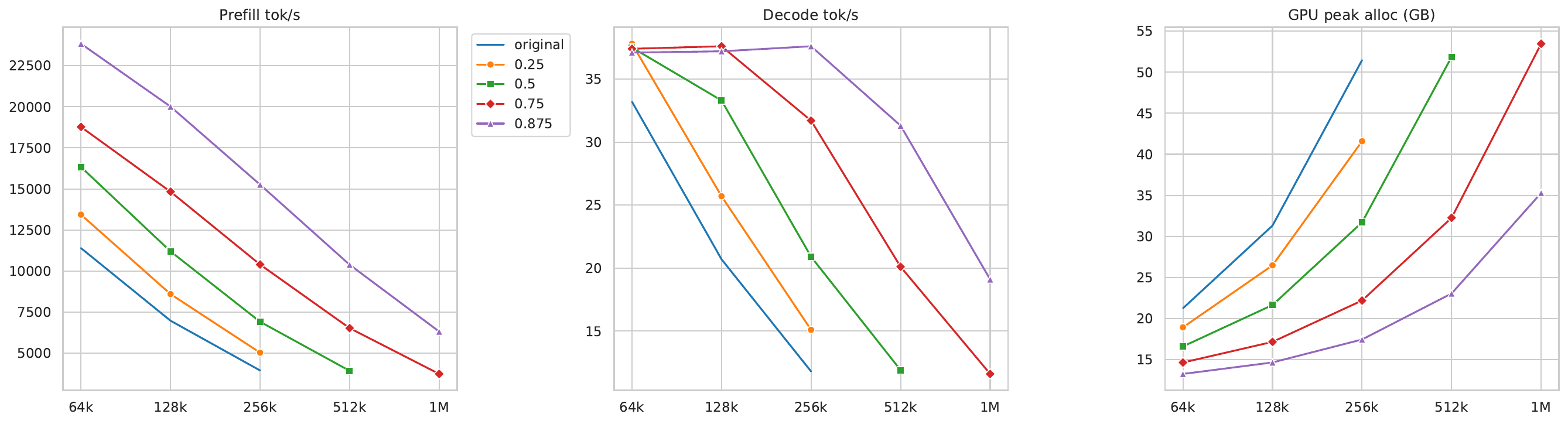}

    \caption{Latency and memory statistics comparing the original Qwen3-8B-Base model with SWA hybrid variants at different SWA $\rho$ ratios using \bosch{}. The left plot shows prefill throughput (tokens/s), the middle plot shows decoding throughput in the same units, and the right plot shows p90 memory allocation (GB). These statistics are measured with input prompts of lengths ranging from 64 to 1M tokens (x-axis). Missing data points correspond to runs where the model encountered out-of-memory (OOM) error. }
    \label{fig:app_latency_qwen3_8b_rho}
\end{figure*}

\autoref{fig:app_latency_qwen3_8b_rho} shows latency and memory usage when comparing the original Qwen3-8B-Base model with \bosch{} SWA hybrid models with $\rho\!\in\!\{0.25,0.5,0.75,0.875\}$ and a window size of 1024. Specifically, we measure prefill and decoding throughput (tokens/s) and the p90 peak memory usage (in GB) when running inference on sequences of varying length (64k, 256k, 512k, and 1M tokens). We run systematic experiments for all models with a batch size of 1 and a streaming context size of 8192 to avoid early out-of-memory (OOM) errors on ultra-long sequences. Missing data points (e.g., for lengths >256k in the original model) indicate that the model ran out of memory (OOM).

\begin{figure*}[!th]
    \centering
    \includegraphics[width=\linewidth]{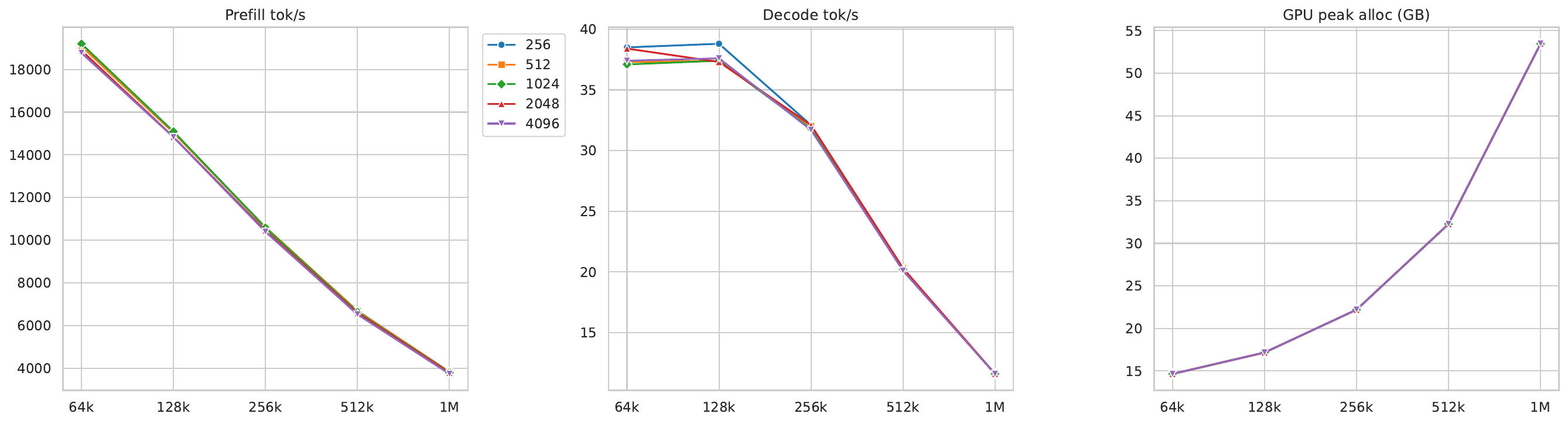}

    \caption{Latency and memory statistics of \bosch{} $\rho\!=\!0.75$ Qwen3-8B-Base SWA hybrid models when varying the SWA window size between 256 and 4096. Notation is the same as in \autoref{fig:app_latency_qwen3_8b_rho}.}
    \label{fig:app_latency_qwen3_8b_ws}
\end{figure*}

Although $\rho\!=\!0.25$ leads to no significant performance drop, it also does not provide efficiency gains: at this ratio, the model still goes OOM beyond 256k tokens, just like the original model despite delivering better latency. In contrast, $\rho\!=\!0.5$ yields improved latency with a moderate performance drop, but it is still not suitable for ultra-long contexts beyond 512k tokens, unlike $\rho\!=\!0.75$ and $\rho\!=\!0.875$, which can handle longer sequences. 

Considering the performance gap between the latter two, $\rho\!=\!0.75$ appears to offer a better trade-off between performance and efficiency. \autoref{fig:app_latency_qwen3_8b_ws} presents the same statistics as \autoref{fig:app_latency_qwen3_8b_rho}, but for the \bosch{} Qwen3-8B-Base SWA hybrid model with $\rho\!=\!0.75$ while varying the SWA window size across $\{256, 512, 1024, 2048, 4096\}$. Despite minor variations for short sequences, we observe that all metrics are almost identical across window sizes when scaling to long sequences. This is mainly because the computational complexity of SWA is effectively constant with respect to the window size, making sequence length the primary driver of complexity. This is encouraging, as it indicates that one can use larger window sizes to benefit from performance gains (see \autoref{app:SWA Windows Size Ablation}) without sacrificing latency.

\section{Results}
\label{app:Results}

\begin{table*}[ht] 
\centering
\resizebox{\textwidth}{!}{

\begin{tabular}{lcccccccccccccccccccccccccccccccccccc}
\toprule
\multirow{3}{*}{\textbf{Method}} &
\multicolumn{16}{c}{\textbf{single}} &
\multicolumn{16}{c}{\textbf{multikey}} &
\multicolumn{4}{c}{\multirow{2}{*}{\textbf{Avg.}}} \\
\cmidrule(lr){2-17}\cmidrule(lr){18-33}
 & \multicolumn{4}{c}{\bf 4k} & \multicolumn{4}{c}{\bf 8k} & \multicolumn{4}{c}{\bf 16k} & \multicolumn{4}{c}{\bf 32k}
 & \multicolumn{4}{c}{\bf 4k} & \multicolumn{4}{c}{\bf 8k} & \multicolumn{4}{c}{\bf 16k} & \multicolumn{4}{c}{\bf 32k}
 & \multicolumn{4}{c}{} \\
\cmidrule(lr){2-5}\cmidrule(lr){6-9}\cmidrule(lr){10-13}\cmidrule(lr){14-17}%
\cmidrule(lr){18-21}\cmidrule(lr){22-25}\cmidrule(lr){26-29}\cmidrule(lr){30-33}%
\cmidrule(lr){34-37}
 & \bf 0.25 & \bf 0.5 & \bf 0.75 & \bf 0.875
 & \bf 0.25 & \bf 0.5 & \bf 0.75 & \bf 0.875
 & \bf 0.25 & \bf 0.5 & \bf 0.75 & \bf 0.875
 & \bf 0.25 & \bf 0.5 & \bf 0.75 & \bf 0.875
 & \bf 0.25 & \bf 0.5 & \bf 0.75 & \bf 0.875
 & \bf 0.25 & \bf 0.5 & \bf 0.75 & \bf 0.875
 & \bf 0.25 & \bf 0.5 & \bf 0.75 & \bf 0.875
 & \bf 0.25 & \bf 0.5 & \bf 0.75 & \bf 0.875
 & \bf 0.25 & \bf 0.5 & \bf 0.75 & \bf 0.875 \\
\midrule
Original & \multicolumn{4}{c}{100.0}  & \multicolumn{4}{c}{99.9} & \multicolumn{4}{c}{100.0} & \multicolumn{4}{c}{100.0} & \multicolumn{4}{c}{98.9} & \multicolumn{4}{c}{94.9} & \multicolumn{4}{c}{84.1} & \multicolumn{4}{c}{63.9} & \multicolumn{4}{c}{92.7}  \\
\midrule
\multicolumn{37}{c}{\textit{\bf Layer-level Heuristics}} \\\midrule
RAND & 95.2 & 71.9 & 32.4 & 32.0 & 87.5 & 58.8 & 15.7 & 15.6 & 76.5 & 22.3 & 6.5 & 7.5 & 65.4 & 21.4 & 2.9 & 2.8 & 71.3 & 51.7 & 19.9 & 17.8 & 47.1 & 23.8 & 11.1 & 10.9 & 32.9 & 14.5 & 4.6 & 4.5 & 23.0 & 7.6 & 1.5 & 1.5 & 66.9 & 38.0 & 12.7 & 12.6 \\
BME  & 85.9 & 32.2 & 32.2 & 32.2 & 69.7 & 15.6 & 15.5 & 15.5 & 60.0 & 7.5 & 7.5 & 7.5 & 43.9 & 2.7 & 2.8 & 2.8 & 33.7 & 15.9 & 19.5 & 19.9 & 16.1 & 9.6 & 10.4 & 10.5 & 11.1 & 4.1 & 3.7 & 4.3 & 5.3 & 1.3 & 1.1 & 1.3 & 40.0 & 11.5 & 11.1 & 11.2 \\
\bottomrule
\end{tabular}
}
\caption{Detailed zero-shot scores on the NIAH benchmark for SWA hybridization methods for Qwen3-1.7B-Base under 4 SWA $\rho$ ratios. The highest score under each configuration is highlighted in bold.}
\label{tab:app_1.7b_niah}
\end{table*}

\begin{table*}[ht]
\centering
\resizebox{\textwidth}{!}{
\begin{tabular}{lcccccccccccccccccccccccccccccccc}
\toprule
\multirow{2}{*}{\textbf{Method}} &
\multicolumn{4}{c}{\bf Single-document QA} &
\multicolumn{4}{c}{\bf Multi-document QA} &
\multicolumn{4}{c}{\bf Summarization} &
\multicolumn{4}{c}{\bf Few-shot learning} &
\multicolumn{4}{c}{\bf Synthetic Tasks} &
\multicolumn{4}{c}{\bf Code Completion} &
\multicolumn{4}{c}{\bf Math Reasoning} &
\multicolumn{4}{c}{\multirow{1}{*}{\textbf{Avg.}}} \\
\cmidrule(lr){2-5}\cmidrule(lr){6-9}\cmidrule(lr){10-13}\cmidrule(lr){14-17}%
\cmidrule(lr){18-21}\cmidrule(lr){22-25}\cmidrule(lr){26-29}\cmidrule(lr){30-33}
 & \bf 0.25 & \bf 0.5 & \bf 0.75 & \bf 0.875
 & \bf 0.25 & \bf 0.5 & \bf 0.75 & \bf 0.875
 & \bf 0.25 & \bf 0.5 & \bf 0.75 & \bf 0.875
 & \bf 0.25 & \bf 0.5 & \bf 0.75 & \bf 0.875
 & \bf 0.25 & \bf 0.5 & \bf 0.75 & \bf 0.875
 & \bf 0.25 & \bf 0.5 & \bf 0.75 & \bf 0.875
 & \bf 0.25 & \bf 0.5 & \bf 0.75 & \bf 0.875
 & \bf 0.25 & \bf 0.5 & \bf 0.75 & \bf 0.875 \\
\midrule
Original
 & \multicolumn{4}{c}{39.8}  & \multicolumn{4}{c}{30.9}  & \multicolumn{4}{c}{26.0}  & \multicolumn{4}{c}{67.4}  & \multicolumn{4}{c}{32.7}
 & \multicolumn{4}{c}{48.2} & \multicolumn{4}{c}{72.6} & \multicolumn{4}{c}{45.4}\\
\midrule
\multicolumn{33}{c}{\textit{\bf Layer-level Heuristics}} \\
\midrule
RAND
 & 27.0 & 17.6 & 13.1 & 14.9
 & 15.7 & 10.0 & 9.4  & 11.1
 & 23.3 & 19.1 & 16.3 & 16.4
 & 56.1 & 46.3 & 36.6 & 36.4
 & 4.2  & 3.6  & 2.6  & 3.2
 & 49.8 & 48.5 & \textbf{47.2} & 44.9
 & 60.0 & \textbf{54.4} & 38.1 & 34.5
 & 33.7 & 28.5 & 23.3 & 23.1 \\
BME
 & 23.1 & 12.7 & 12.9 & 14.1
 & 13.8 & 11.3 & 11.1 & 10.2
 & 21.1 & 14.8 & 15.7 & 15.9
 & 58.1 & 36.8 & 37.8 & 37.0
 & 3.2  & 2.9  & 2.8  & 2.5
 & 50.0 & 32.3 & 34.6 & 36.7
 & 51.0 & 36.8 & 29.9 & 30.3
 & 31.5 & 21.1 & 20.7 & 20.9 \\
\bottomrule
\end{tabular}
}
\caption{Detailed zero-shot scores on the LongBench benchmark for SWA hybridization methods for Qwen3-1.7B-Base under 4 SWA $\rho$ ratios. The highest score under each configuration is highlighted in bold.}
\label{tab:app_1.7b_longbench}
\end{table*}

\begin{table*}[ht] 
\centering
\resizebox{\textwidth}{!}{
\begin{tabular}{lcccccccccccccccccccccccccccccccccccc}
\toprule
\multirow{3}{*}{\textbf{Method}} &
\multicolumn{16}{c}{\textbf{single}} &
\multicolumn{16}{c}{\textbf{multikey}} &
\multicolumn{4}{c}{\multirow{2}{*}{\textbf{Avg.}}} \\
\cmidrule(lr){2-17}\cmidrule(lr){18-33}
 & \multicolumn{4}{c}{\bf 4k} & \multicolumn{4}{c}{\bf 8k} & \multicolumn{4}{c}{\bf 16k} & \multicolumn{4}{c}{\bf 32k}
 & \multicolumn{4}{c}{\bf 4k} & \multicolumn{4}{c}{\bf 8k} & \multicolumn{4}{c}{\bf 16k} & \multicolumn{4}{c}{\bf 32k}
 & \multicolumn{4}{c}{} \\
\cmidrule(lr){2-5}\cmidrule(lr){6-9}\cmidrule(lr){10-13}\cmidrule(lr){14-17}%
\cmidrule(lr){18-21}\cmidrule(lr){22-25}\cmidrule(lr){26-29}\cmidrule(lr){30-33}%
\cmidrule(lr){34-37}
 & \bf 0.25 & \bf 0.5 & \bf 0.75 & \bf 0.875
 & \bf 0.25 & \bf 0.5 & \bf 0.75 & \bf 0.875
 & \bf 0.25 & \bf 0.5 & \bf 0.75 & \bf 0.875
 & \bf 0.25 & \bf 0.5 & \bf 0.75 & \bf 0.875
 & \bf 0.25 & \bf 0.5 & \bf 0.75 & \bf 0.875
 & \bf 0.25 & \bf 0.5 & \bf 0.75 & \bf 0.875
 & \bf 0.25 & \bf 0.5 & \bf 0.75 & \bf 0.875
 & \bf 0.25 & \bf 0.5 & \bf 0.75 & \bf 0.875
 & \bf 0.25 & \bf 0.5 & \bf 0.75 & \bf 0.875 \\
\midrule
Original
 & \multicolumn{4}{c}{100.0}  & \multicolumn{4}{c}{100.0}  & \multicolumn{4}{c}{100.0}  & \multicolumn{4}{c}{99.9}
 & \multicolumn{4}{c}{99.5} & \multicolumn{4}{c}{99.6} & \multicolumn{4}{c}{99.1} & \multicolumn{4}{c}{94.5}
 & \multicolumn{4}{c}{99.1} \\
\midrule
\multicolumn{37}{c}{\textit{\bf Layer-level Heuristics}} \\
\midrule
RAND
& 68.5 & 37.0 & 32.0 & 32.2 & 60.4 & 17.6 & 15.8 & 15.6 & 53.6 & 8.1 & 7.5 & 7.5 & 49.9 & 3.3 & 2.8 & 2.8 & 58.8 & 34.6 & 26.5 & 27.6 & 38.5 & 14.4 & 11.7 & 12.8 & 22.7 & 6.0 & 4.5 & 5.3 & 15.0 & 2.1 & 1.6 & 1.7 & 45.9 & 15.4 & 12.8 & 13.2 \\
\bottomrule
\end{tabular}
}
\caption{Detailed zero-shot scores on the NIAH benchmark for SWA hybridization methods for Qwen3-8B-Base under 4 SWA $\rho$ ratios. The highest score under each configuration is highlighted in bold.}
\label{tab:app_8b_niah}
\end{table*}

\begin{table*}[ht]
\centering
\resizebox{\textwidth}{!}{
\begin{tabular}{lcccccccccccccccccccccccccccccccc}
\toprule
\multirow{2}{*}{\textbf{Method}} &
\multicolumn{4}{c}{\bf Single-document QA} &
\multicolumn{4}{c}{\bf Multi-document QA} &
\multicolumn{4}{c}{\bf Summarization} &
\multicolumn{4}{c}{\bf Few-shot learning} &
\multicolumn{4}{c}{\bf Synthetic Tasks} &
\multicolumn{4}{c}{\bf Code Completion} &
\multicolumn{4}{c}{\bf Math Reasoning} &
\multicolumn{4}{c}{\multirow{1}{*}{\textbf{Avg.}}} \\
\cmidrule(lr){2-5}\cmidrule(lr){6-9}\cmidrule(lr){10-13}\cmidrule(lr){14-17}%
\cmidrule(lr){18-21}\cmidrule(lr){22-25}\cmidrule(lr){26-29}\cmidrule(lr){30-33}
 & \bf 0.25 & \bf 0.5 & \bf 0.75 & \bf 0.875
 & \bf 0.25 & \bf 0.5 & \bf 0.75 & \bf 0.875
 & \bf 0.25 & \bf 0.5 & \bf 0.75 & \bf 0.875
 & \bf 0.25 & \bf 0.5 & \bf 0.75 & \bf 0.875
 & \bf 0.25 & \bf 0.5 & \bf 0.75 & \bf 0.875
 & \bf 0.25 & \bf 0.5 & \bf 0.75 & \bf 0.875
 & \bf 0.25 & \bf 0.5 & \bf 0.75 & \bf 0.875
 & \bf 0.25 & \bf 0.5 & \bf 0.75 & \bf 0.875 \\
\midrule
Original
& \multicolumn{4}{c}{45.5} & \multicolumn{4}{c}{41.9} & \multicolumn{4}{c}{26.8} & \multicolumn{4}{c}{70.3}
& \multicolumn{4}{c}{63.0} & \multicolumn{4}{c}{66.6} & \multicolumn{4}{c}{85.8}
& \multicolumn{4}{c}{57.1} \\
\midrule
\multicolumn{33}{c}{\textit{\bf Layer-level Heuristics}} \\
\midrule
RAND & 25.8 & 14.6 & 7.5 & 9.8 & 20.4 & 11.6 & 5.3 & 7.4 & 24.5 & 18.7 & 16.6 & 16.5 & 62.9 & 50.0 & 38.1 & 36.5 & 12.8 & 3.8 & 3.5 & 3.0 & 52.5 & 43.2 & 32.9 & 33.1 & 82.3 & 78.8 & 65.2 & 64.0 & 40.2 & 31.5 & 24.1 & 24.3 \\
BME & 19.3 & 10.8 & 14.4 & 10.3 & 15.4 & 9.8 & 14.8 & 12.0 & 21.0 & 19.2 & 20.6 & 18.8 & 50.7 & 42.6 & 41.4 & 38.2 & 4.0 & 3.8 & 2.9 & 2.7 & 29.9 & 26.3 & 45.0 & 42.0 & 76.3 & 58.2 & 62.9 & 61.3 & 31.0 & 24.4 & 28.9 & 26.5 \\
\bottomrule
\end{tabular}
}
\caption{Detailed zero-shot scores on the LongBench benchmark for SWA hybridization methods for Qwen3-8B-Base under 4 SWA $\rho$ ratios. The highest score under each configuration is highlighted in bold.}
\label{tab:app_8b_longbench}
\end{table*}

\begin{table*}[ht]
\centering
\resizebox{\textwidth}{!}{
\begin{tabular}{lcccccccccccccccccccccccccccccccccccc}
\toprule
\multirow{3}{*}{\textbf{Method}} &
\multicolumn{16}{c}{\textbf{single}} &
\multicolumn{16}{c}{\textbf{multikey}} &
\multicolumn{4}{c}{\multirow{2}{*}{\textbf{Avg.}}} \\
\cmidrule(lr){2-17}\cmidrule(lr){18-33}
 & \multicolumn{4}{c}{\bf 4k} & \multicolumn{4}{c}{\bf 8k} & \multicolumn{4}{c}{\bf 16k} & \multicolumn{4}{c}{\bf 32k}
 & \multicolumn{4}{c}{\bf 4k} & \multicolumn{4}{c}{\bf 8k} & \multicolumn{4}{c}{\bf 16k} & \multicolumn{4}{c}{\bf 32k}
 & \multicolumn{4}{c}{} \\
\cmidrule(lr){2-5}\cmidrule(lr){6-9}\cmidrule(lr){10-13}\cmidrule(lr){14-17}%
\cmidrule(lr){18-21}\cmidrule(lr){22-25}\cmidrule(lr){26-29}\cmidrule(lr){30-33}%
\cmidrule(lr){34-37}
 & \bf 0.25 & \bf 0.5 & \bf 0.75 & \bf 0.875
 & \bf 0.25 & \bf 0.5 & \bf 0.75 & \bf 0.875
 & \bf 0.25 & \bf 0.5 & \bf 0.75 & \bf 0.875
 & \bf 0.25 & \bf 0.5 & \bf 0.75 & \bf 0.875
 & \bf 0.25 & \bf 0.5 & \bf 0.75 & \bf 0.875
 & \bf 0.25 & \bf 0.5 & \bf 0.75 & \bf 0.875
 & \bf 0.25 & \bf 0.5 & \bf 0.75 & \bf 0.875
 & \bf 0.25 & \bf 0.5 & \bf 0.75 & \bf 0.875
 & \bf 0.25 & \bf 0.5 & \bf 0.75 & \bf 0.875 \\
\midrule
Original
& \multicolumn{4}{c}{100.0} & \multicolumn{4}{c}{99.9} & \multicolumn{4}{c}{100.0} & \multicolumn{4}{c}{100.0} & \multicolumn{4}{c}{99.8} & \multicolumn{4}{c}{99.7} & \multicolumn{4}{c}{99.3} & \multicolumn{4}{c}{98.2} & \multicolumn{4}{c}{99.6} \\
\midrule
\multicolumn{37}{c}{\textit{\bf Layer-level Heuristics}} \\
\midrule
RAND
 & 99.9 & 86.4 & 42.1 & 32.3
 & 97.3 & 59.1 & 22.8 & 15.6
 & 92.9 & 47.8 & 11.7 & 7.5
 & 82.9 & 31.7 & 8.8 & 2.8
 & 96.2 & 54.0 & 30.7 & 26.9
 & 74.6 & 22.4 & 13.5 & 12.5
 & 53.0 & 9.4 & 5.5 & 4.9
 & 35.6 & 3.0 & 1.6 & 1.6
 & 79.0 & 39.2 & 17.1 & 13.0 \\
BME
 & 98.5 & 31.3 & 32.3 & 32.3
 & 82.5 & 15.5 & 15.6 & 15.6
 & 43.7 & 7.1 & 7.4 & 7.5
 & 34.1 & 2.7 & 2.8 & 2.8
 & 41.3 & 22.1 & 20.6 & 21.9
 & 20.8 & 11.1 & 11.0 & 11.4
 & 7.1 & 3.7 & 4.2 & 4.6
 & 2.1 & 1.3 & 1.6 & 1.5
 & 41.3 & 11.8 & 11.9 & 12.2 \\
\bottomrule
\end{tabular}
}
\caption{Detailed zero-shot scores on the NIAH benchmark for SWA hybridization methods for Qwen3-14B-Base under 4 SWA $\rho$ ratios. The highest score under each configuration is highlighted in bold.}
\label{tab:app_14b_niah}
\end{table*}

\begin{table*}[ht]
\centering
\resizebox{\textwidth}{!}{
\begin{tabular}{lcccccccccccccccccccccccccccccccc}
\toprule
\multirow{2}{*}{\textbf{Method}} &
\multicolumn{4}{c}{\bf Single-document QA} &
\multicolumn{4}{c}{\bf Multi-document QA} &
\multicolumn{4}{c}{\bf Summarization} &
\multicolumn{4}{c}{\bf Few-shot learning} &
\multicolumn{4}{c}{\bf Synthetic Tasks} &
\multicolumn{4}{c}{\bf Code Completion} &
\multicolumn{4}{c}{\bf Math Reasoning} &
\multicolumn{4}{c}{\multirow{1}{*}{\textbf{Avg.}}} \\%
\cmidrule(lr){2-5}\cmidrule(lr){6-9}\cmidrule(lr){10-13}\cmidrule(lr){14-17}%
\cmidrule(lr){18-21}\cmidrule(lr){22-25}\cmidrule(lr){26-29}\cmidrule(lr){30-33}
 & \bf 0.25 & \bf 0.5 & \bf 0.75 & \bf 0.875
 & \bf 0.25 & \bf 0.5 & \bf 0.75 & \bf 0.875
 & \bf 0.25 & \bf 0.5 & \bf 0.75 & \bf 0.875
 & \bf 0.25 & \bf 0.5 & \bf 0.75 & \bf 0.875
 & \bf 0.25 & \bf 0.5 & \bf 0.75 & \bf 0.875
 & \bf 0.25 & \bf 0.5 & \bf 0.75 & \bf 0.875
 & \bf 0.25 & \bf 0.5 & \bf 0.75 & \bf 0.875
 & \bf 0.25 & \bf 0.5 & \bf 0.75 & \bf 0.875 \\ 
\midrule
Original
 & \multicolumn{4}{c}{43.7}  
 & \multicolumn{4}{c}{43.5}  
 & \multicolumn{4}{c}{27.0}  
 & \multicolumn{4}{c}{72.1}  
 & \multicolumn{4}{c}{66.7}  
 & \multicolumn{4}{c}{66.6}  
 & \multicolumn{4}{c}{89.2}  
 & \multicolumn{4}{c}{58.4}  
 \\ 
\midrule
\multicolumn{33}{c}{\textit{\bf Layer-level Heuristics}} \\
\midrule
RAND
& 32.1 & 21.9 & 17.6 & 15.8
& 26.7 & 17.1 & 15.2 & 14.7
& 26.3 & 24.2 & 22.4 & 21.6
& 68.0 & 58.6 & 52.0 & 49.8
& 42.5 & 9.8 & 4.0 & 3.7
& 58.2 & \textbf{56.4} & 55.0 & 57.0
& 84.0 & 75.7 & 63.8 & 57.0
& 48.2 & 37.6 & 32.8 & 31.4 \\

BME
& 25.1 & 5.3 & 15.8 & 14.6
& 23.2 & 3.5 & 10.8 & 11.7
& 18.8 & 11.6 & 21.4 & 21.7
& 47.9 & 18.5 & 46.4 & 50.4
& 3.5  & 2.3 & 3.7  & 3.8
& 41.2 & 25.4 & 50.0 & 53.9
& 68.9 & 45.0 & 55.2 & 55.3
& 32.6 & 15.9 & 29.1 & 30.2 \\
\bottomrule
\end{tabular}
}
\caption{Detailed zero-shot scores on the LongBench benchmark for SWA hybridization methods for Qwen3-14B-Base under 4 SWA $\rho$ ratios. The highest score under each configuration is highlighted in bold.}
\label{tab:app_14b_longbench}
\end{table*}

\begin{table*}[ht]
\centering
\resizebox{\textwidth}{!}{
\begin{tabular}{lcccccccccccccccccccccccccccccccccccc}
\toprule
\multirow{3}{*}{\textbf{Method}} &
\multicolumn{16}{c}{\textbf{single}} &
\multicolumn{16}{c}{\textbf{multikey}} &
\multicolumn{4}{c}{\multirow{2}{*}{\textbf{Avg.}}} \\
\cmidrule(lr){2-17}\cmidrule(lr){18-33}
& \multicolumn{4}{c}{\bf 4k} & \multicolumn{4}{c}{\bf 8k} & \multicolumn{4}{c}{\bf 16k} & \multicolumn{4}{c}{\bf 32k}
& \multicolumn{4}{c}{\bf 4k} & \multicolumn{4}{c}{\bf 8k} & \multicolumn{4}{c}{\bf 16k} & \multicolumn{4}{c}{\bf 32k}
& \multicolumn{4}{c}{} \\
\cmidrule(lr){2-5}\cmidrule(lr){6-9}\cmidrule(lr){10-13}\cmidrule(lr){14-17}%
\cmidrule(lr){18-21}\cmidrule(lr){22-25}\cmidrule(lr){26-29}\cmidrule(lr){30-33}%
\cmidrule(lr){34-37}
& \bf 0.25 & \bf 0.5 & \bf 0.75 & \bf 0.875
& \bf 0.25 & \bf 0.5 & \bf 0.75 & \bf 0.875
& \bf 0.25 & \bf 0.5 & \bf 0.75 & \bf 0.875
& \bf 0.25 & \bf 0.5 & \bf 0.75 & \bf 0.875
& \bf 0.25 & \bf 0.5 & \bf 0.75 & \bf 0.875
& \bf 0.25 & \bf 0.5 & \bf 0.75 & \bf 0.875
& \bf 0.25 & \bf 0.5 & \bf 0.75 & \bf 0.875
& \bf 0.25 & \bf 0.5 & \bf 0.75 & \bf 0.875 \\
\midrule
Original
& \multicolumn{4}{c}{100.0} & \multicolumn{4}{c}{100.0} & \multicolumn{4}{c}{99.9} & \multicolumn{4}{c}{100.0}
& \multicolumn{4}{c}{99.7} & \multicolumn{4}{c}{99.6} & \multicolumn{4}{c}{99.1} & \multicolumn{4}{c}{96.7}
& \multicolumn{4}{c}{99.4} \\
\midrule
\multicolumn{37}{c}{\textit{\bf Layer-level Heuristics}} \\
\midrule
RAND & 100.0 & 31.0 & 30.8 & 30.6 & 99.8 & 22.2 & 21.9 & 18.0
& 99.9 & 20.0 & 19.8 & 10.8 & 97.3 & 9.7 & 9.8 & 2.9
& 87.7 & 22.0 & 19.9 & 19.5 & 61.9 & 8.6 & 7.9 & 7.9
& 51.6 & 4.2 & 4.3 & 3.7 & 35.5 & 1.3 & 1.4 & 1.2
& 79.6 & 16.0 & 15.0 & 10.6 \\
BME & 80.1 & 45.7 & 32.3 & 29.7 & 87.9 & 43.3 & 15.3 & 12.8
& 88.1 & 37.9 & 6.4 & 6.2 & 45.2 & 23.2 & 2.8 & 2.8
& 38.5 & 17.9 & 23.8 & 23.5 & 18.3 & 11.6 & 10.2 & 9.9
& 15.9 & 6.7 & 3.6 & 3.9 & 6.2 & 2.3 & 1.3 & 1.5
& 47.5 & 18.9 & 12.0 & 10.5 \\
\bottomrule
\end{tabular}
}
\caption{Detailed zero-shot scores on the NIAH benchmark for SWA hybridization methods for Qwen3-30B-A3B-Base under 4 SWA $\rho$ ratios. The highest score under each configuration is highlighted in bold.}
\label{tab:app_30b_niah}
\end{table*}

\begin{table*}[ht]
\centering
\resizebox{\textwidth}{!}{
\begin{tabular}{lcccccccccccccccccccccccccccccccc}
\toprule
\multirow{2}{*}{\textbf{Method}} &
\multicolumn{4}{c}{\bf Single-document QA} &
\multicolumn{4}{c}{\bf Multi-document QA} &
\multicolumn{4}{c}{\bf Summarization} &
\multicolumn{4}{c}{\bf Few-shot learning} &
\multicolumn{4}{c}{\bf Synthetic Tasks} &
\multicolumn{4}{c}{\bf Code Completion} &
\multicolumn{4}{c}{\bf Math Reasoning} &
\multicolumn{4}{c}{\multirow{1}{*}{\textbf{Avg.}}} \\
\cmidrule(lr){2-5}\cmidrule(lr){6-9}\cmidrule(lr){10-13}\cmidrule(lr){14-17}%
\cmidrule(lr){18-21}\cmidrule(lr){22-25}\cmidrule(lr){26-29}\cmidrule(lr){30-33}
 & \bf 0.25 & \bf 0.5 & \bf 0.75 & \bf 0.875
 & \bf 0.25 & \bf 0.5 & \bf 0.75 & \bf 0.875
 & \bf 0.25 & \bf 0.5 & \bf 0.75 & \bf 0.875
 & \bf 0.25 & \bf 0.5 & \bf 0.75 & \bf 0.875
 & \bf 0.25 & \bf 0.5 & \bf 0.75 & \bf 0.875
 & \bf 0.25 & \bf 0.5 & \bf 0.75 & \bf 0.875
 & \bf 0.25 & \bf 0.5 & \bf 0.75 & \bf 0.875
 & \bf 0.25 & \bf 0.5 & \bf 0.75 & \bf 0.875 \\
\midrule
Original
 & \multicolumn{4}{c}{24.6}
 & \multicolumn{4}{c}{39.8}
 & \multicolumn{4}{c}{27.6}
 & \multicolumn{4}{c}{70.0}
 & \multicolumn{4}{c}{64.3}
 & \multicolumn{4}{c}{67.8}
 & \multicolumn{4}{c}{90.1}
 & \multicolumn{4}{c}{54.9} \\
\midrule
\multicolumn{33}{c}{\textit{\bf Layer-level Heuristics}} \\
\midrule
RAND
 & \textbf{14.3} & 4.6 & 4.0 & 3.2
 & 18.6 & 3.9 & 3.9 & 3.6
 & 26.2 & 11.4 & 8.6 & 7.0
 & 67.7 & 28.1 & 19.5 & 17.8
 & 26.6 & 3.1 & 2.7 & 3.3
 & 58.8 & 31.3 & 25.8 & 25.3
 & 69.4 & 45.6 & 44.7 & 36.4
 & 40.2 & 18.3 & 15.6 & 13.8 \\

BME
 & 9.5 & 4.2 & 4.2 & 3.7
 & 5.9 & 3.9 & 3.8 & 3.7
 & 21.9 & 10.1 & 7.3 & 8.4
 & 55.1 & 28.9 & 25.0 & 20.9
 & 3.1 & 2.7 & 3.2 & 3.0
 & 32.0 & 23.6 & 17.4 & 30.6
 & 70.1 & 43.1 & 25.5 & 27.3
 & 28.2 & 16.7 & 12.3 & 14.0 \\
\bottomrule
\end{tabular}
}
\caption{Detailed zero-shot scores on the LongBench benchmark for SWA hybridization methods for Qwen3-30B-A3B-Base under 4 SWA $\rho$ ratios. The highest score under each configuration is highlighted in bold.}
\label{tab:app_30b_longbench}
\end{table*}

\begin{table*}[ht]
\centering
\resizebox{\textwidth}{!}{
\begin{tabular}{lcccccccccccccccccccccccccccccccccccc}
\toprule
\multirow{3}{*}{\textbf{Method}} &
\multicolumn{16}{c}{\textbf{single}} &
\multicolumn{16}{c}{\textbf{multikey}} &
\multicolumn{4}{c}{\multirow{2}{*}{\textbf{Avg.}}} \\
\cmidrule(lr){2-17}\cmidrule(lr){18-33}
 & \multicolumn{4}{c}{\bf 4k} & \multicolumn{4}{c}{\bf 8k} & \multicolumn{4}{c}{\bf 16k} & \multicolumn{4}{c}{\bf 32k}
 & \multicolumn{4}{c}{\bf 4k} & \multicolumn{4}{c}{\bf 8k} & \multicolumn{4}{c}{\bf 16k} & \multicolumn{4}{c}{\bf 32k}
 & \multicolumn{4}{c}{} \\
\cmidrule(lr){2-5}\cmidrule(lr){6-9}\cmidrule(lr){10-13}\cmidrule(lr){14-17}%
\cmidrule(lr){18-21}\cmidrule(lr){22-25}\cmidrule(lr){26-29}\cmidrule(lr){30-33}%
\cmidrule(lr){34-37}
 & \bf 0.7B & \bf 1.3B & \bf 2.0B & \bf 2.5B
 & \bf 0.7B & \bf 1.3B & \bf 2.0B & \bf 2.5B
 & \bf 0.7B & \bf 1.3B & \bf 2.0B & \bf 2.5B
 & \bf 0.7B & \bf 1.3B & \bf 2.0B & \bf 2.5B
 & \bf 0.7B & \bf 1.3B & \bf 2.0B & \bf 2.5B
 & \bf 0.7B & \bf 1.3B & \bf 2.0B & \bf 2.5B
 & \bf 0.7B & \bf 1.3B & \bf 2.0B & \bf 2.5B
 & \bf 0.7B & \bf 1.3B & \bf 2.0B & \bf 2.5B
 & \bf 0.7B & \bf 1.3B & \bf 2.0B & \bf 2.5B \\
\midrule
\multicolumn{37}{c}{\textit{\bf Qwen3-1.7B-Base}} \\
\midrule
Original
 & \multicolumn{4}{c}{100.0}  & \multicolumn{4}{c}{99.9}  & \multicolumn{4}{c}{100.0}  & \multicolumn{4}{c}{100.0}
 & \multicolumn{4}{c}{98.9} & \multicolumn{4}{c}{94.9} & \multicolumn{4}{c}{84.1} & \multicolumn{4}{c}{63.9}
 & \multicolumn{4}{c}{92.7} \\
\midrule
INTR
 & 95.9 & 96.4 & 96.5 & 95.7
 & 91.6 & 93.4 & 95.1 & 93.6
 & 72.6 & 75.5 & 78.7 & 76.7
 & 52.2 & 57.3 & 59.4 & 56.4
 & 80.3 & 87.3 & 88.4 & 88.9
 & 64.1 & 74.9 & 78.9 & 79.1
 & 36.4 & 49.5 & 56.9 & 57.3
 & 17.9 & 24.7 & 30.1 & 31.7
 & 63.9 & 69.9 & 73.0 & 72.4 \\
\bottomrule
\end{tabular}
}
\caption{Detailed continual pretraining performances on the NIAH benchmark for 4 SWA hybridization methods for 4 Qwen3 models (1.7B, 8B, 14B, 30B) under SWA ratio $\rho\!=\!0.75$. The highest score under each configuration is highlighted in bold.}
\label{tab:app_sft_niah}
\end{table*}

\begin{table*}[ht]
\centering
\resizebox{\textwidth}{!}{
\begin{tabular}{lcccccccccccccccccccccccccccccccc}
\toprule
\multirow{2}{*}{\textbf{Method}} &
\multicolumn{4}{c}{\bf Single-document QA} &
\multicolumn{4}{c}{\bf Multi-document QA} &
\multicolumn{4}{c}{\bf Summarization} &
\multicolumn{4}{c}{\bf Few-shot learning} &
\multicolumn{4}{c}{\bf Synthetic Tasks} &
\multicolumn{4}{c}{\bf Code Completion} &
\multicolumn{4}{c}{\bf Math Reasoning} &
\multicolumn{4}{c}{\multirow{1}{*}{\textbf{Avg.}}} \\
\cmidrule(lr){2-5}\cmidrule(lr){6-9}\cmidrule(lr){10-13}\cmidrule(lr){14-17}%
\cmidrule(lr){18-21}\cmidrule(lr){22-25}\cmidrule(lr){26-29}\cmidrule(lr){30-33}
 & \bf 0.7B & \bf 1.3B & \bf 2.0B & \bf 2.5B
 & \bf 0.7B & \bf 1.3B & \bf 2.0B & \bf 2.5B
 & \bf 0.7B & \bf 1.3B & \bf 2.0B & \bf 2.5B
 & \bf 0.7B & \bf 1.3B & \bf 2.0B & \bf 2.5B
 & \bf 0.7B & \bf 1.3B & \bf 2.0B & \bf 2.5B
 & \bf 0.7B & \bf 1.3B & \bf 2.0B & \bf 2.5B
 & \bf 0.7B & \bf 1.3B & \bf 2.0B & \bf 2.5B
 & \bf 0.7B & \bf 1.3B & \bf 2.0B & \bf 2.5B \\
\midrule
\multicolumn{33}{c}{\textit{\bf Qwen3-1.7B-Base}} \\
\midrule
Original
 & \multicolumn{4}{c}{39.8}  & \multicolumn{4}{c}{30.9}  & \multicolumn{4}{c}{26.0}  & \multicolumn{4}{c}{67.4}  & \multicolumn{4}{c}{32.7}
 & \multicolumn{4}{c}{48.2} & \multicolumn{4}{c}{72.6} & \multicolumn{4}{c}{45.4}
\\
\midrule
INTR
 & 25.9 & 28.1 & 28.4 & 29.4
 & 22.6 & 23.8 & 25.3 & 25.2
 & 20.6 & 21.1 & 21.3 & 21.5
 & 55.6 & 55.8 & 57.0 & 57.9
 & 3.8 & 4.7 & 4.7 & 4.7
 & 50.8 & 47.3 & 46.8 & 45.8
 & 58.7 & 59.9 & 60.6 & 61.1
 & 34.0 & 34.4 & 34.9 & 35.1 \\
\textit{B-layer}
 & \textbf{35.4} & 36.5 & 37.3 & 37.2
 & 25.4 & 27.4 & \textbf{29.4} & \textbf{30.3}
 & \textbf{21.6} & 22.4 & 22.8 & 22.6
 & 60.8 & 62.4 & 62.6 & 63.0
 & 6.0 & 10.0 & \textbf{13.0} & \textbf{13.2}
 & 48.2 & 46.9 & 46.1 & 44.2
 & 57.8 & 60.3 & 59.8 & 61.1
 & 36.9 & 38.4 & 38.7 & 38.8 \\
\bottomrule
\end{tabular}
}
\caption{Detailed continual pretraining performances on the LongBench benchmark for 4 SWA hybridization methods for 4 Qwen3 models (1.7B, 8B, 14B, 30B) under SWA ratio $\rho\!=\!0.75$. The highest score under each configuration is highlighted in bold.}
\label{tab:app_sft_niah}
\end{table*}

\begin{table*}[ht]
\centering
\resizebox{\textwidth}{!}{
\begin{tabular}{lcccccccccccccccccccccccccccccccccccc}
\toprule
\multirow{3}{*}{\textbf{Method}} &
\multicolumn{16}{c}{\textbf{single}} &
\multicolumn{16}{c}{\textbf{multikey}} &
\multicolumn{4}{c}{\multirow{2}{*}{\textbf{Avg.}}} \\
\cmidrule(lr){2-17}\cmidrule(lr){18-33}
 & \multicolumn{4}{c}{\bf 4k} & \multicolumn{4}{c}{\bf 8k} & \multicolumn{4}{c}{\bf 16k} & \multicolumn{4}{c}{\bf 32k}
 & \multicolumn{4}{c}{\bf 4k} & \multicolumn{4}{c}{\bf 8k} & \multicolumn{4}{c}{\bf 16k} & \multicolumn{4}{c}{\bf 32k}
 & \multicolumn{4}{c}{} \\
\cmidrule(lr){2-5}\cmidrule(lr){6-9}\cmidrule(lr){10-13}\cmidrule(lr){14-17}%
\cmidrule(lr){18-21}\cmidrule(lr){22-25}\cmidrule(lr){26-29}\cmidrule(lr){30-33}%
\cmidrule(lr){34-37}
 & \bf 256 & \bf 512 & \bf 2k & \bf 4k
 & \bf 256 & \bf 512 & \bf 2k & \bf 4k
 & \bf 256 & \bf 512 & \bf 2k & \bf 4k
 & \bf 256 & \bf 512 & \bf 2k & \bf 4k
 & \bf 256 & \bf 512 & \bf 2k & \bf 4k
 & \bf 256 & \bf 512 & \bf 2k & \bf 4k
 & \bf 256 & \bf 512 & \bf 2k & \bf 4k
 & \bf 256 & \bf 512 & \bf 2k & \bf 4k
 & \bf 256 & \bf 512 & \bf 2k & \bf 4k
  \\
\midrule

\multicolumn{37}{c}{\textit{\bf Qwen3-1.7B-Base}} \\
\midrule
Original
 & \multicolumn{4}{c}{100.0}  & \multicolumn{4}{c}{99.9}  & \multicolumn{4}{c}{100.0}  & \multicolumn{4}{c}{100.0}
 & \multicolumn{4}{c}{98.9} & \multicolumn{4}{c}{94.9} & \multicolumn{4}{c}{84.1} & \multicolumn{4}{c}{63.9}
 & \multicolumn{4}{c}{92.7} \\
\midrule
INTR
 & 5.1 & 13.8 & 55.6 & \textbf{100.0} & 3.2 & 7.3 & 30.9 & 55.5 & 2.4
 & 2.7 & 14.2 & 28.9 & 2.1 & 2.3 & 6.8 & 13.1 & 4.1 & 9.5
 & 40.7 & \textbf{98.9} & 2.6 & 6.1 & 16.3 & 27.6 & 1.6 & 2.3 & 7.9
 & 13.5 & 0.9 & 1.2 & 3.9 & 5.5 & 2.8 & 5.7 & 22.0 & 42.9 \\
\bottomrule
\end{tabular}
}
\caption{Detailed zero-shot performances on the NIAH benchmark for 4 SWA hybridization methods for 4 Qwen3 models (1.7B, 8B, 14B, 30B) under SWA ratio $\rho\!=\!0.75$ when SWA windows size is $\{256,512,2048,4096\}$. The highest score under each configuration is highlighted in bold.}
\label{tab:app_ws_niah_075}
\end{table*}

\begin{table*}[ht]
\centering
\resizebox{\textwidth}{!}{
\begin{tabular}{lcccccccccccccccccccccccccccccccc}
\toprule
\multirow{2}{*}{\textbf{Method}} &
\multicolumn{4}{c}{\bf Single-document QA} &
\multicolumn{4}{c}{\bf Multi-document QA} &
\multicolumn{4}{c}{\bf Summarization} &
\multicolumn{4}{c}{\bf Few-shot learning} &
\multicolumn{4}{c}{\bf Synthetic Tasks} &
\multicolumn{4}{c}{\bf Code Completion} &
\multicolumn{4}{c}{\bf Math Reasoning} &
\multicolumn{4}{c}{\multirow{1}{*}{\textbf{Avg.}}} \\
\cmidrule(lr){2-5}\cmidrule(lr){6-9}\cmidrule(lr){10-13}\cmidrule(lr){14-17}%
\cmidrule(lr){18-21}\cmidrule(lr){22-25}\cmidrule(lr){26-29}\cmidrule(lr){30-33}
 & \bf 256 & \bf 512 & \bf 2k & \bf 4k
 & \bf 256 & \bf 512 & \bf 2k & \bf 4k
 & \bf 256 & \bf 512 & \bf 2k & \bf 4k
 & \bf 256 & \bf 512 & \bf 2k & \bf 4k
 & \bf 256 & \bf 512 & \bf 2k & \bf 4k
 & \bf 256 & \bf 512 & \bf 2k & \bf 4k
 & \bf 256 & \bf 512 & \bf 2k & \bf 4k
 & \bf 256 & \bf 512 & \bf 2k & \bf 4k \\
\midrule
\multicolumn{33}{c}{\textit{\bf Qwen3-1.7B-Base}} \\
\midrule
Original
 & \multicolumn{4}{c}{39.8}  & \multicolumn{4}{c}{30.9}  & \multicolumn{4}{c}{26.0}  & \multicolumn{4}{c}{67.4}  & \multicolumn{4}{c}{32.7}
 & \multicolumn{4}{c}{48.2} & \multicolumn{4}{c}{72.6} & \multicolumn{4}{c}{45.4}
\\
\midrule

INTR
 & 13.2 & 13.9 & 16.8 & 25.1
 & \textbf{11.4} & \textbf{11.9} & 12.6 & 13.3
 & 11.8 & 14.6 & 18.3 & 20.0
 & 19.2 & 27.9 & 47.1 & 55.8
 & \textbf{3.1} & 3.5 & \textbf{3.8} & 4.9
 & 24.6 & 23.9 & 30.4 & 39.0
 & 3.9 & 4.6 & 71.0 & 71.0
 & 12.5 & 14.3 & 28.6 & 32.7 \\

\textit{B-layer}
 & 13.3 & \textbf{15.8} & 21.6 & 27.9
 & 8.9 & 10.9 & 13.7 & 16.0
 & 14.0 & \textbf{18.0} & \textbf{21.7} & \textbf{23.1}
 & 29.2 & 34.6 & 49.5 & 55.3
 & 3.0 & \textbf{3.7} & 3.6 & \textbf{5.8}
 & 31.2 & 32.2 & 37.2 & 43.6
 & \textbf{15.2} & \textbf{19.6} & 71.0 & 71.0
 & 16.4 & \textbf{19.3} & 31.2 & 34.7 \\
\bottomrule
\end{tabular}
}
\caption{Detailed zero-shot performances on the LongBench benchmark for 4 SWA hybridization methods for 4 Qwen3 models (1.7B, 8B, 14B, 30B) under SWA ratio $\rho\!=\!0.75$ when SWA windows size is $\{256,512,2048,4096\}$. The highest score under each configuration is highlighted in bold.}
\label{tab:app_ws_longbench_075}
\end{table*}

\begin{table*}[ht]
\centering
\begin{tabular}{lccccccccc}
\toprule
\multirow{2}{*}{\textbf{Method}} &
\multicolumn{4}{c}{\textbf{single}} &
\multicolumn{4}{c}{\textbf{multikey}} &
\multirow{2}{*}{\textbf{Avg.}} \\
\cmidrule(lr){2-5}\cmidrule(lr){6-9}
 & \bf 4k & \bf 8k & \bf 16k & \bf 32k
 & \bf 4k & \bf 8k & \bf 16k & \bf 32k
 & \\
\midrule
\multicolumn{10}{c}{\textit{\bf Qwen3-1.7B-Base}} \\
\midrule
$0.25 \rightarrow 0.5$  & 81.6 & 80.4 & 71.0 & 70.2 & 51.9 & 41.5 & 31.3 & 29.1 & 57.1 \\
$0.5 \rightarrow 0.75$  & 69.8 & 61.0 & 52.9 & 45.3 & 53.6 & 44.5 & 29.7 & 20.0 & 47.1 \\
$0.75 \rightarrow 0.875$ & 44.8 & 31.5 & 13.1 & 10.7 & 22.8 & 18.7 &  8.0 &  8.0 & 19.7 \\

\midrule
\multicolumn{10}{c}{\textit{\bf Qwen3-8B-Base}} \\
\midrule
$0.25 \rightarrow 0.5$  & 96.9 & 94.9 & 93.8 & 80.9 & 91.2 & 77.4 & 65.3 & 35.7 & 79.5 \\
$0.5 \rightarrow 0.75$  & 93.3 & 91.1 & 87.0 & 70.6 & 80.3 & 59.8 & 43.3 & 23.8 & 68.7 \\
$0.75 \rightarrow 0.875$ & 44.6 & 31.7 & 21.4 & 13.0 & 31.0 & 14.9 &  6.6 &  2.0 & 20.7 \\

\midrule
\multicolumn{10}{c}{\textit{\bf Qwen3-14B-Base}} \\
\midrule
$0.25 \rightarrow 0.5$  & 99.9 & 97.6 & 95.4 & 83.9 & 97.9 & 91.9 & 82.4 & 53.8 & 87.9 \\
$0.5 \rightarrow 0.75$  & 92.9 & 79.2 & 93.9 & 65.4 & 92.9 & 68.0 & 59.0 & 40.1 & 73.9 \\
$0.75 \rightarrow 0.875$ & 68.8 & 49.9 & 52.4 & 27.3 & 59.6 & 32.6 & 25.4 & 12.5 & 41.1 \\

\midrule
\multicolumn{10}{c}{\textit{\bf Qwen3-30B-A3B-Base}} \\
\midrule
$0.25 \rightarrow 0.5$  & 94.9 & 91.6 & 76.7 & 59.3 & 89.2 & 64.0 & 36.8 & 24.9 & 67.2 \\
$0.5 \rightarrow 0.75$  & 69.1 & 61.0 & 52.2 & 30.7 & 41.8 & 22.8 & 12.5 &  4.0 & 36.8 \\
$0.75 \rightarrow 0.875$ & 35.9 & 26.6 & 21.5 &  9.2 & 20.8 &  7.5 &  3.4 &  0.8 & 15.7 \\

\bottomrule
\end{tabular}
\caption{Zero-shot scores on the NIAH  benchmark for \bosch{} when using $\mathcal{A}^\prime_{\rho_l}$ across 4 Qwen3 models  under 3 SWA ratios $\rho\in\{0.5,0.75,0.875\}$. The left side of $\rightarrow$ corresponds to $\rho_s$, while the right side corresponds to $\rho_l$. $\mathcal{A}^\prime_{\rho_l}$ indicates results obtained after replacing a randomly selected subset of heads at a given configuration (e.g., $\rho_l\!=\!0.75$) with heads that did not appear in a smaller configuration (e.g., $\rho_s\!=\!0.5$). Scores are averaged over three runs with different seeds.}

\label{tab:app_turnover_niah}
\end{table*}

\begin{table*}[ht]
\centering
\resizebox{\textwidth}{!}{
\begin{tabular}{lcccccccc}
\toprule
\multirow{2}{*}{\textbf{Method}} &
\multicolumn{7}{c}{\textbf{Tasks}} &
\multirow{2}{*}{\textbf{Avg.}} \\
\cmidrule(lr){2-8}
 & \textbf{Single-document QA}
 & \textbf{Multi-document QA}
 & \textbf{Summarization}
 & \textbf{Few-shot learning}
 & \textbf{Synthetic Tasks}
 & \textbf{Code Completion}
 & \textbf{Math Reasoning}
 & \\
\midrule
\multicolumn{9}{c}{\textit{\bf Qwen3-1.7B-Base}} \\
\midrule
$0.25 \rightarrow 0.5$
& 15.0 & 8.9 & 18.3 & 48.3 & 3.8 & 47.4 & 49.8 & 27.4 \\
$0.5 \rightarrow 0.75$
& 11.6 & 9.1 & 17.1 & 43.4 & 2.9 & 38.4 & 38.8 & 23.0 \\
$0.75 \rightarrow 0.875$
& 4.8 & 3.2 & 12.9 & 39.9 & 2.2 & 24.0 & 31.6 & 17.0 \\

\midrule
\multicolumn{9}{c}{\textit{\bf Qwen3-8B-Base}} \\
\midrule
$0.25 \rightarrow 0.5$
& 22.2 & 16.1 & 22.4 & 58.4 & 10.2 & 42.6 & 78.1 & 34.3 \\
$0.5 \rightarrow 0.75$
& 16.4 & 12.7 & 21.2 & 47.3 & 5.1 & 33.4 & 72.7 & 29.2 \\
$0.75 \rightarrow 0.875$
& 13.0 & 5.5 & 15.8 & 43.0 & 3.2 & 33.0 & 64.9 & 24.3 \\

\midrule
\multicolumn{9}{c}{\textit{\bf Qwen3-14B-Base}} \\
\midrule
$0.25 \rightarrow 0.5$
& 24.6 & 28.0 & 24.6 & 65.1 & 24.7 & 52.5 & 84.1 & 40.4 \\
$0.5 \rightarrow 0.75$
& 18.0 & 14.7 & 21.7 & 55.6 & 13.1 & 52.9 & 80.2 & 34.0 \\
$0.75 \rightarrow 0.875$
& 14.2 & 11.0 & 17.2 & 49.4 & 4.2 & 50.3 & 72.0 & 29.9 \\

\midrule
\multicolumn{9}{c}{\textit{\bf Qwen3-30B-A3B-Base}} \\
\midrule
$0.25 \rightarrow 0.5$
& 9.0 & 5.4 & 21.4 & 54.2 & 2.7 & 56.6 & 70.9 & 28.4 \\
$0.5 \rightarrow 0.75$
& 5.8 & 5.6 & 14.3 & 40.6 & 3.3 & 34.5 & 54.8 & 22.3 \\
$0.75 \rightarrow 0.875$
& 3.3 & 3.2 & 6.6 & 23.0 & 3.7 & 23.7 & 31.8 & 14.7 \\

\bottomrule
\end{tabular}
}
\caption{Zero-shot scores on the LongBench  benchmark for \bosch{} when using $\mathcal{A}^\prime_{\rho_l}$ across 4 Qwen3 models  under 3 SWA ratios $\rho\in\{0.5,0.75,0.875\}$ (right side of $\rightarrow$). $\mathcal{A}^\prime_{\rho_l}$ indicates results obtained after replacing a randomly selected subset of heads at a given configuration (e.g., $\rho_l\!=\!0.75$) with heads that did not appear in a smaller configuration (e.g., $\rho_s\!=\!0.5$). Scores are averaged over three runs with different seeds.}
\label{tab:app_turnover_longbench}
\end{table*}

\begin{figure*}[b]
    \centering
    \includegraphics[height=0.4\textheight]{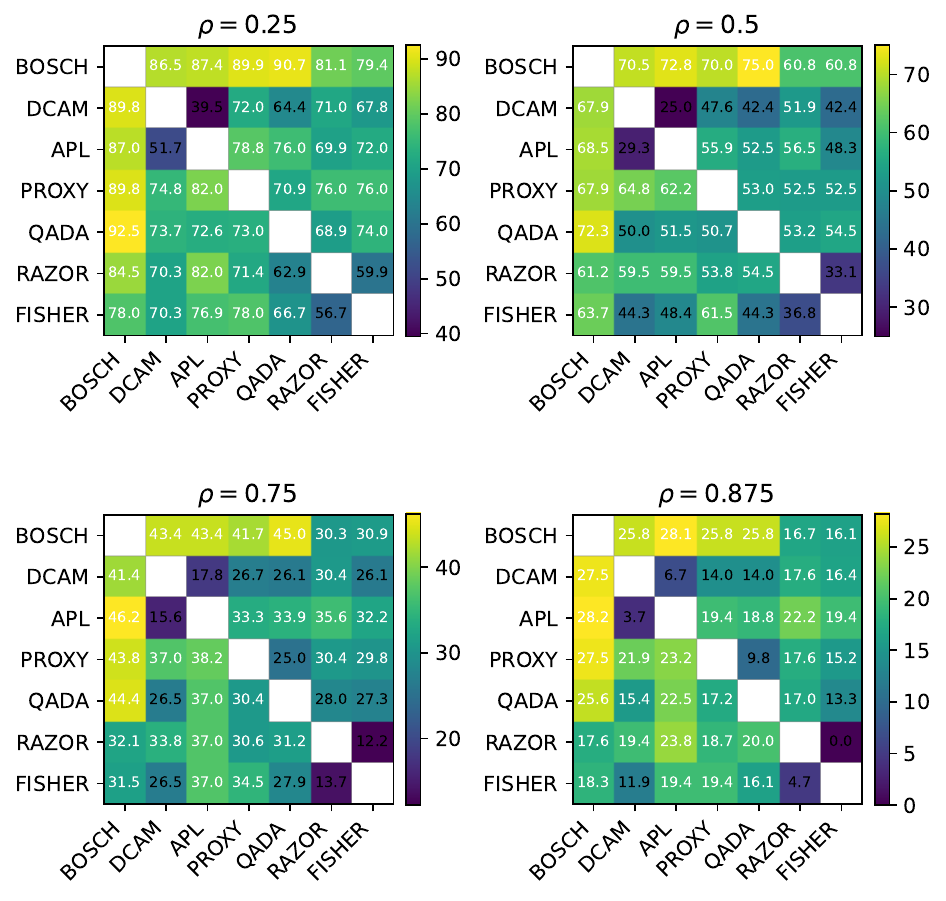}
    \caption{Jaccard distance between SWA heads selected by seven methods for $\rho\in \{0.25,0.5,0.75,0.875\}$. The distances for Qwen3-8B-Base  and Qwen3-14B-Base models are shown in the lower and upper triangles, respectively.}
    \label{fig:heatmap_app_8b_14b}
\end{figure*}

\begin{figure*}[b]
    \centering
    \includegraphics[height=0.4\textheight]{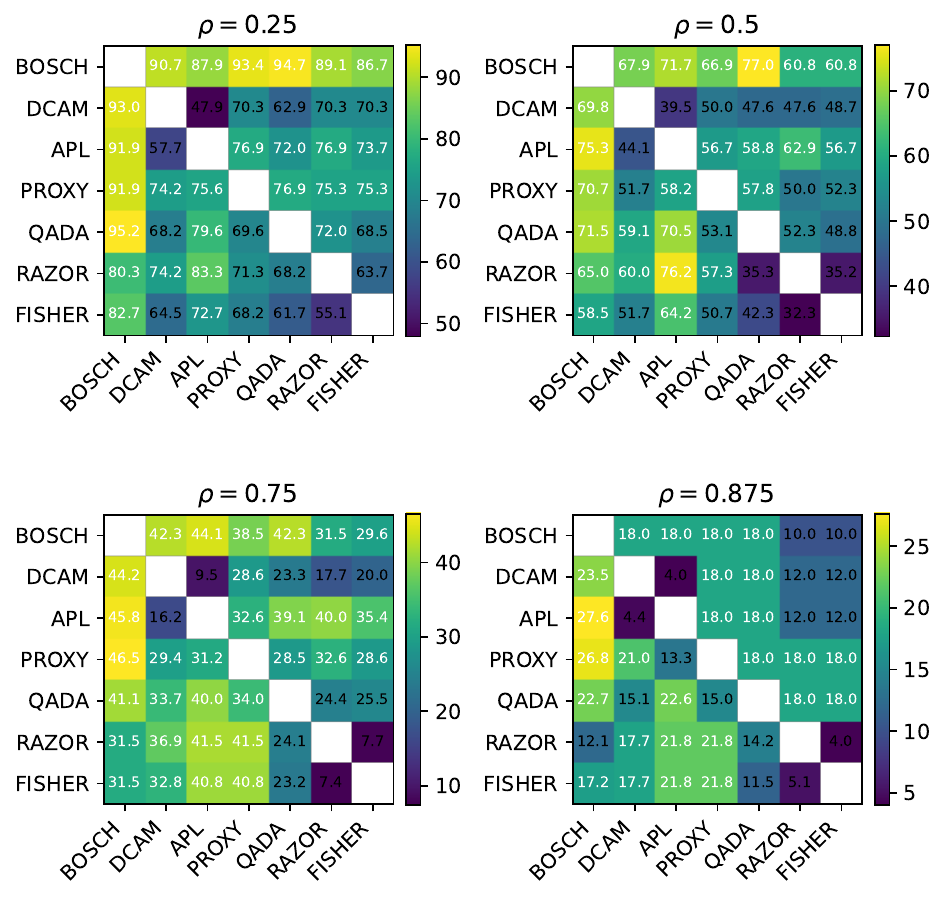}
    \caption{Jaccard distance between SWA heads selected by seven methods for $\rho\in \{0.25,0.5,0.75,0.875\}$. The distances for Qwen3-1.7B-Base and Qwen3-30B-A3B-Base models are shown in the lower and upper triangles, respectively.}
    \label{fig:heatmap_app_1.7b_30b}
\end{figure*}

\end{document}